\RequirePackage{fix-cm}
\documentclass[smallcondensed]{svjour3}     
\smartqed  
\usepackage{times}

\usepackage[utf8]{inputenc}
\usepackage[english]{babel}

\usepackage[numbers]{natbib}
\usepackage{multicol}
\usepackage{array}
\usepackage{makecell}
\usepackage{multirow}
\usepackage{graphicx}
\usepackage{tabularx}
\usepackage{amsmath}
\usepackage{amssymb}
\usepackage[T1]{fontenc}
\usepackage[linesnumbered,lined,ruled,commentsnumbered]{algorithm2e}
\usepackage[small]{caption}
\usepackage{graphicx}
\usepackage{subcaption}
\usepackage{mathtools}
\usepackage{blkarray}
\usepackage[dvipsnames]{xcolor}
\usepackage[colorlinks=true,linkcolor=black,anchorcolor=black,citecolor=black,filecolor=black,menucolor=black,runcolor=black,urlcolor=black]{hyperref}
\usepackage[utf8]{inputenc}
\usepackage[english]{babel}

\DeclareMathOperator*{\argmax}{arg\,max}
\DeclareMathOperator{\Tr}{Tr}
\usepackage[shortlabels]{enumitem}
\newtheorem{prop}{Proposition}
\usepackage{mathtools}
\DeclarePairedDelimiter{\ceil}{\lceil}{\rceil}
\usepackage{scalerel,stackengine}
\stackMath
\newcommand\reallywidehat[1]{%
	\savestack{\tmpbox}{\stretchto{%
			\scaleto{%
				\scalerel*[\widthof{\ensuremath{#1}}]{\kern-.6pt\bigwedge\kern-.6pt}%
				{\rule[-\textheight/2]{1ex}{\textheight}}
			}{\textheight}%
		}{0.5ex}}%
	\stackon[1pt]{#1}{\tmpbox}%
}

\begin{document}

\title{Multi-UAV Planning for Cooperative Wildfire Coverage and Tracking with Quality-of-Service Guarantees\thanks{\textbf{To appear in the journal of Autonomous Agents and Multi-Agent Systems (AAMAS)}. This work was sponsored by ONR under grant N00014-18-S-B001, MIT Lincoln Laboratory grant 7000437192, Lockheed Martin Corporation under grant GR00000509, and GaTech institute funding.}
}

\titlerunning{Coordinated Planning and Collaborative Field Coverage}        

\author{Esmaeil Seraj         \and
	    Andrew Silva         \and
        Matthew Gombolay 
}


\institute{E. Seraj \at
              Georgia Institute of Technology, Atlanta, GA, 30332, USA \\
              \email{eseraj3@gatech.edu}           
           \and
           A. Silva \at
           Georgia Institute of Technology, Atlanta, GA, 30332, USA \\
           \email{andrew.silva@gatech.edu}
           \and
           M. Gombolay \at
              Georgia Institute of Technology, Atlanta, GA, 30332, USA \\
              \email{matthew.gombolay@cc.gatech.edu}
}

\date{Received: 05.15.2022 / Accepted: 05.24.2022}

\maketitle

\begin{abstract}
In recent years, teams of robot and Unmanned Aerial Vehicles (UAVs) have been commissioned by researchers to enable accurate, online wildfire coverage and tracking. While the majority of prior work focuses on the coordination and control of such multi-robot systems, to date, these UAV teams have not been given the ability to reason about a fire's track (i.e., location and propagation dynamics) to provide performance guarantee over a time horizon. Motivated by the problem of aerial wildfire monitoring, we propose a predictive framework which enables cooperation in multi-UAV teams towards collaborative field coverage and fire tracking with probabilistic performance guarantee. Our approach enables UAVs to infer the latent fire propagation dynamics for time-extended coordination in safety-critical conditions. We derive a set of novel, analytical temporal, and tracking-error bounds to enable the UAV-team to distribute their limited resources and cover the entire fire area according to the case-specific estimated states and provide a probabilistic performance guarantee. Our results are not limited to the aerial wildfire monitoring case-study and are generally applicable to problems, such as search-and-rescue, target tracking and border patrol. We evaluate our approach in simulation and provide demonstrations of the proposed framework on a physical multi-robot testbed to account for real robot dynamics and restrictions. Our quantitative evaluations validate the performance of our method accumulating $7.5\times$ and $9.0\times$ smaller tracking-error than state-of-the-art model-based and reinforcement learning benchmarks, respectively.
\end{abstract}

\section{Introduction}
\label{sec:introduction}
\noindent \textcolor{black}{Unmanned Aerial Vehicles (UAVs) have gained significant interest \textcolor{black}{across} research communities due to their agility, cost-efficiency, \textcolor{black}{and ability to access and monitor remote areas.} Multi-robot systems, such as teams of UAVs, have been commissioned as Mobile Sensor Networks (MSNs) in applications which require observing (i.e., monitoring and surveilling) the physical world and gathering information~\cite{caillouet2018optimization, zhao2019systemic}. MSNs have numerous applications in surveillance and disaster response~\cite{akyildiz2004wireless, liu2011last, afghah2019wildfire, doherty2009temporal}, environment monitoring and agriculture~\cite{li2008nonthreshold, ota2012oracle}, wildfire fighting~\cite{seraj2020coordinated, pham2017distributed, pham2018distributed}, search-and-rescue~\cite{liu2011last}, manufacturing~\cite{bays2015solution, mozaffari2016efficient}, homeland security and border patrol~\cite{ahmadzadeh2006cooperative, xia2007wireless}.}

While multi-robot systems are capable of executing time-sensitive, complex missions that are intractable for a single robot, it is challenging to efficiently coordinate such systems and to optimize the collaborative behavior among robots~\cite{seraj2021hierarchical, konan2022iterated}. This coordinated planning problem becomes even more challenging when robots have to collaborate \textcolor{black}{in a dynamic environment, such as in monitoring an active wildfire. Dynamic environments are restless; meaning regardless of robots' collective actions, the states of the environment continually change. As such, classical multi-agent planning and scheduling approaches designed for static environments will fail in such domains, since coordination plans made for a timestep may be inapplicable to the next step due to the environment dynamicity~\cite{brenner2009continual, undeger2010multi}.}

In this paper, we study the problem of coordinated planning of multi-UAV teams deployed as MSNs \textcolor{black}{for cooperative surveillance and tracking of a dynamic environment. In our work, we enable UAVs with the ability to estimate the changing states of their environment and leverage these estimated information to reason about their cooperation plan for collaborative monitoring. Particularly, we consider safety-critical and time-sensitive scenarios where only a limited number of UAVs are available. Therefore, in our problem, not only it is important to reason robustly under environmental dynamicity via active planning with limited resources, it is exigent to have probabilistic guarantees that the UAV team can execute the coordinated plan. In our approach, the UAV agents actively evaluate a probabilistic error-bound that provides such guarantee and attempt to revise their coordinated plan when a \textcolor{black}{performance guarantee} cannot be provided due to changes in environment states. While the problem of multi-agent planning and scheduling has been \textcolor{black}{studied extensively} in prior work~\cite{brenner2009continual, undeger2010multi, ravichandar2020strata, sarne2013determining, panait2005cooperative, prasad1999learning}, many of these works either lack the ability to actively plan in a dynamic environment, or do not provide a probabilistic \textcolor{black}{performance guarantee} that assures the teams' success in executing a coordinated plan.}

\subsection{\textcolor{black}{Motivating Application: Aerial Wildfire Monitoring}}
\label{subsec:MotivatingApplicationAerialWildfireMonitoring}
\noindent We \textcolor{black}{adopt} the application of aerial wildfire monitoring as a running case-study and motivate our paper in this \textcolor{black}{important safety-critical problem.} In the application of wildfire fighting, human firefighters on the ground need online and dynamic observation of the firefront (\textcolor{black}{i.e.}, the moving edge of fire) to anticipate a wildfire’s unknown characteristics, such as size, scale, and propagation velocity, in order to plan their strategies accordingly. \textcolor{black}{To support human firefighters,} teams of UAVs can be deployed as MSNs to estimate the states of fire across thousands of acres and provide human firefighters with such information~\cite{de2020aerial, xing2019kalman, afghah2019wildfire}. Nevertheless, coordinated control of UAV teams in this volatile setting provides particular challenges~\cite{beard2006decentralized,mcintire2016iterated,nunes2015multi,choi2009consensus,allison2016airborne}. 

Fighting wildfires safely and effectively requires accurate online information on firefront location, size, shape, and propagation velocity~\cite{martinez2008computer,stipanivcev2010advanced,sujit2007cooperative,de2020aerial}. To provide firefighters with this realtime information, researchers have sought to utilize satellite feeds to estimate fire location information~\cite{fujiwara2002forest,kudoh2003two,casbeer2006cooperative}. Unfortunately, the resolution of these images is too low for more than simple detection of a wildfire's existence~\cite{casbeer2006cooperative}. Firefighters need frequent, high-quality images of the wildfire to make strategic plans~\cite{de2020aerial,pham2017distributed,haksar2018distributed}. Here, we define \textit{high-quality information} as local, high-resolution images (or other sensory information) that are captured from a close by distance with respect to the areas prioritized by humans.
\begin{figure}[t!]
	\centering
	\begin{subfigure}[t]{0.55\linewidth}
		\centering
		\includegraphics[width=\linewidth]{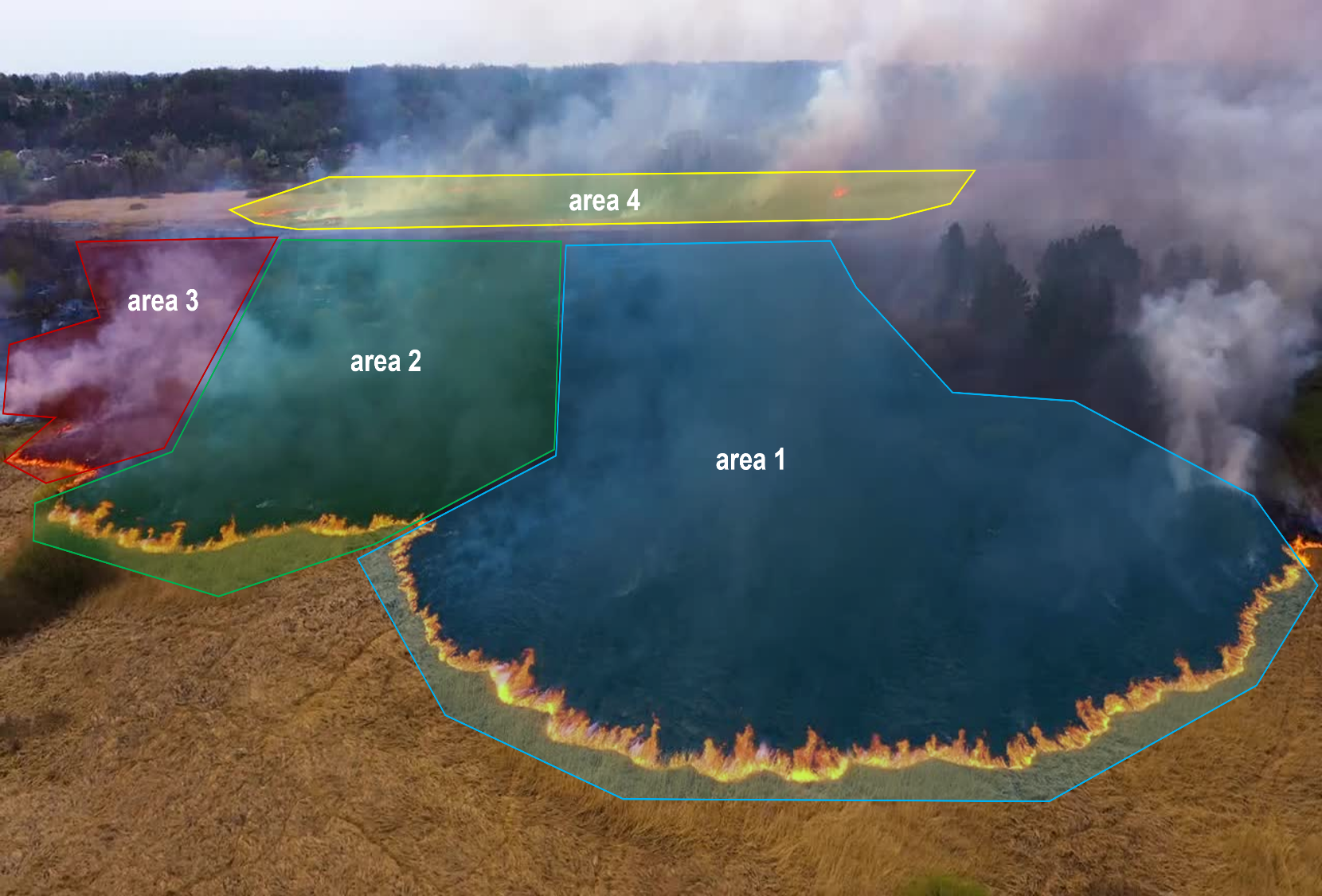}
		\caption{\textcolor{black}{Clustering fire areas for monitoring based on propagation velocity and direction \textcolor{black}{(courtesy of Motion Array; used with modifications)}.}}
		\label{fig:AreaCovergae}
	\end{subfigure}
	~~
	\begin{subfigure}[t]{0.41\linewidth}
		\centering
		\includegraphics[width=\linewidth]{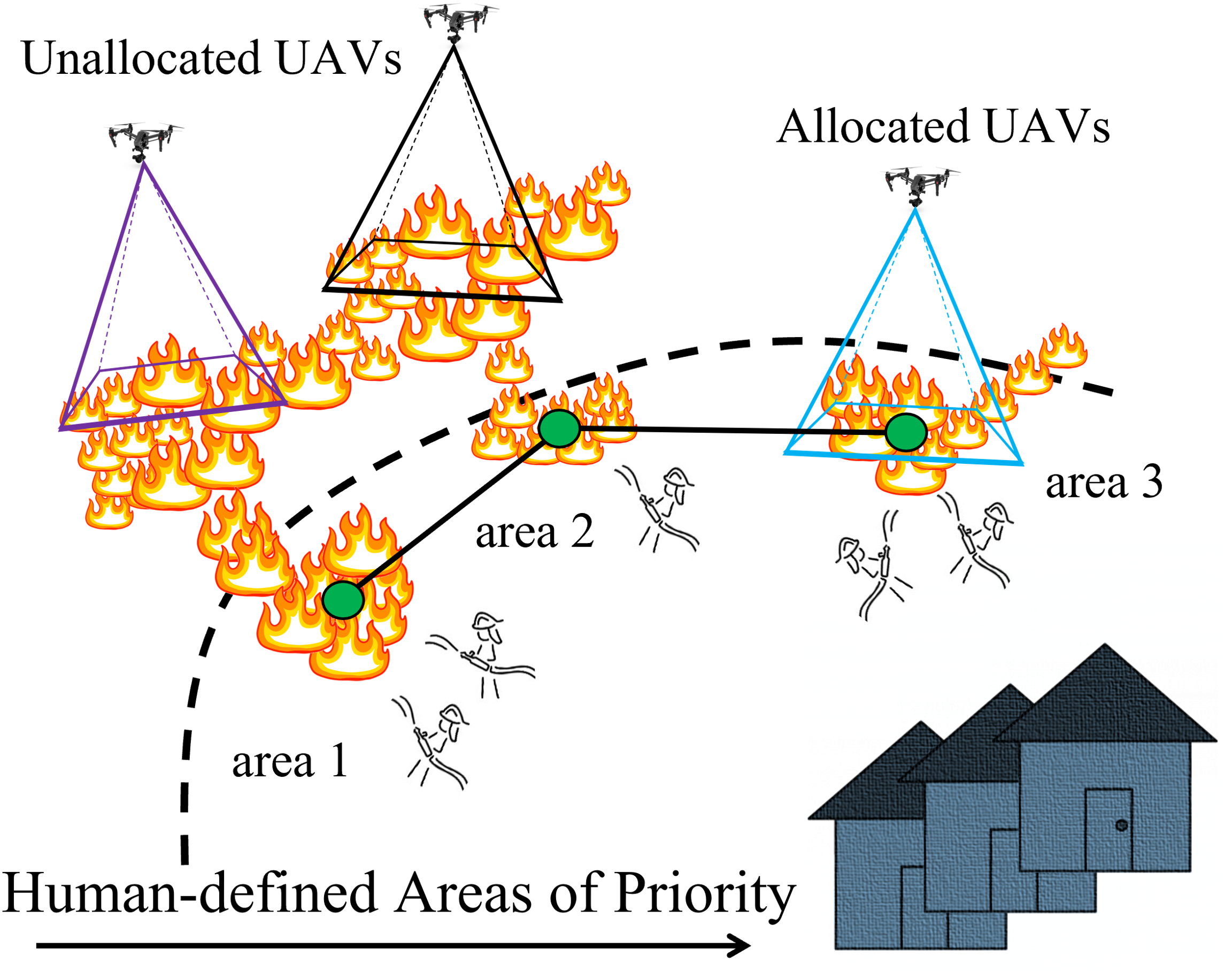}
		\caption{\textcolor{black}{Separating fire areas for coverage and tracking based on areas of human activity and priority.}}
		\label{fig:SafeHRI}
	\end{subfigure}
	\caption{Figure~\ref{fig:AreaCovergae} demonstrates an example of separating wildfire areas for coverage and tracking based on fire propagation velocity and direction. The firefront in each area is moving in a different direction and with a varying velocity and, therefore, at least one separate UAV is required to monitor and track the firefront in each area. In a similar perspective, \textcolor{black}{Figure~\ref{fig:SafeHRI}} shows an example of disjoint coverage areas that are separated according to areas of human (\textcolor{black}{i.e.}, firefighters) activity and priority. In this case, the autonomous UAVs are teaming with humans to safely manage their human collaborators by providing for them \textit{high-quality} information regarding their time-varying proximity to fire (i.e., areas of human operation). \textcolor{black}{When we have access to additional UAV resources that may not be needed in the prioritized areas, we can deploy such UAVs (i.e., unallocated UAVs) to monitor the rest of the wildfire areas (not specified or prioritized).}}
	\vspace*{-0.5cm}
\end{figure}

\textcolor{black}{In large scale dynamic field coverage and tracking applications, such as aerial wildfire monitoring, UAVs need to be distributed effectively to cover the entire area of interest~\cite{seraj2020coordinated, pham2018distributed}. In many cases, the region that need to be covered and tracked by the UAV team can be sub-divided into smaller areas according to different characteristics, priorities or needs. Figure~\ref{fig:AreaCovergae} depicts an example of separating wildfire areas for coverage and tracking based on fire propagation velocity and direction. The firefront in each area is moving in a different direction and with a varying velocity and therefore, at least one separate UAV is required to monitor and track the firefront in each area. In a similar perspective, \textcolor{black}{Figure~\ref{fig:SafeHRI}} shows an example of disjoint coverage areas that are separated according to regions of human (\textcolor{black}{i.e.}, firefighters) activity and priority. In this case, the autonomous UAVs are teaming with humans to safely manage their human collaborators by providing for them \textit{high-quality} information regarding their time-varying proximity to fire (i.e., areas of human operation). Accordingly, when the UAV resources are limited, a multi-agent planning algorithm is required such that the real-time cooperative field coverage and tracking performance can be guaranteed with as few UAV agents as possible.} 

\subsection{\textcolor{black}{Contributions}}
\label{subsec:Contributions}
\noindent \textcolor{black}{To overcome the aforementioned limitations, herein we develop a probabilistically-\textcolor{black}{guaranteed} framework for real-time, large-scale coordinated planning and cooperative coverage \textcolor{black}{of dynamic environments}. We \textcolor{black}{ground} our \textcolor{black}{approach} \textcolor{black}{in \textcolor{black}{application to aerial wildfire} monitoring and develop an} algorithmic framework in which we explicitly infer the latent fire propagation dynamics and leverage the estimated information in real-time to plan for a guaranteed \textcolor{black}{quality of service.} In our predictive mechanism, we define the term ``service'' as actively estimating the states of a propagating wildfire via limited number of UAVs and providing human firefighters on the ground with probabilistic guarantees regarding their proximity to a fire. We design our planning and coverage (\textcolor{black}{i.e.}, monitoring) strategies upon} a multi-step Adaptive Extended Kalman Filter (AEKF) predictor and the FARSITE wildfire propagation mathematical model~\cite{finney1998farsite}.

A primary motivation of our approach is the need for computationally lightweight, yet probabilistically-\textcolor{black}{guaranteed}, algorithms to scale to large numbers of UAVs and coverage \textcolor{black}{areas}. A central contribution of our work is a set of novel, analytical temporal and tracking-error residual bounds that allow UAVs to enable probabilistically-\textcolor{black}{guaranteed coordination in monitoring and tracking dynamic targets (i.e., firespots)}. We derive these bounds in three different scenarios: (1) stationary fire, (2) moving fire without considerable grow (i.e., spawn), and (3) quickly moving, multiplying fire. For each case, we derive \textcolor{black}{analytic and probabilistic temporal upper-bounds to facilitate \textcolor{black}{performance guarantee} for collaborative field coverage}. The primary contributions of our work are:
\begin{enumerate}
    \item An algorithmic \textcolor{black}{planning} framework based on Kalman estimation and uncertainty propagation to learn and leverage in real-time a predictive mechanism that enables probabilistic guarantees for tracking dynamic \textcolor{black}{targets (i.e., firefronts)}.
    
    \item \textcolor{black}{Introducing the Uncertainty Residual Ratio (URR) tracking-error bound as well as} a set of novel, analytical temporal upper-bounds that allow UAVs to enable probabilistically-\textcolor{black}{guaranteed coordination in monitoring and tracking dynamic targets (i.e., firespots)}
    
    \item An efficient and scalable collaborative field coverage algorithm by leveraging our analytical temporal bounds for centralized planning and distributed execution.
    
    \item \textcolor{black}{Presenting quantitative and experimental results in simulation and on physical robots that show the effectiveness of our URR bound and cooperative coordinated multi-UAV planning framework by demonstrating efficient scalability and significantly outperforming two prior SOTA baselines~\cite{pham2017distributed, haksar2018distributed} in a wildfire coverage and tracking task.}
\end{enumerate} 

\noindent We note that, despite the utility of UAVs as the robot agents and the aerial wildfire monitoring application in our paper, the proposed coordinated, collaborative planning and area coverage algorithms as well as the derived set of analytical temporal \textcolor{black}{and the URR} tracking-error bounds are broadly applicable to other \textcolor{black}{safety-critical applications that require mobile sensors to accurately track moving targets in their environment. See Section~\ref{subsubsec:ApplicabilityofProposedAlgorithms} for a detailed discussion of the applicability of our proposed framework.} \textcolor{black}{We investigate the cooperative field monitoring problem by considering the wildfires in} three separate scenarios of \textit{stationary}, \textit{moving} and \textit{moving-spreading} target points \textcolor{black}{(e.g., firespots).}

In the application of wildfire fighting, human teams (\textcolor{black}{i.e.}, firefighters) fight a wildfire until the fire is extinguished or working condition becomes too harsh to continue relative to the cost of \textcolor{black}{abandoning}. However, determining when a situation is too dangerous relative to the costs of abandoning the effort is a non-trivial ethical dilemma (e.g., how to measure the lives of immovable patients at a hospital in the fire's path relative to those of the firefighters) complicated by a lack of information regarding the fire's propagation characteristics. We take the perspective that this decision should remain in the hands of human professionals. As such, in our case-study, we focus 1) on developing a tight, probabilistic, upper-bound \textcolor{black}{to reason about the minimum} number of robots required to ensure high-quality information for the human decision-makers (Section \ref{sec:method}) and 2) measuring the tightness of our test by the number of robots required to satisfy this bound relative to other state-of-the-art approaches (Section \ref{sec:experimentalresults}).

We empirically evaluate the performance, feasibility, and scalability of our framework in various experiments, alongside a state-of-the-art model-based~\cite{pham2017distributed} and a reinforcement learning (RL) benchmark~\cite{haksar2018distributed} for UAV-based aerial field coverage and coordinated planning. Our experiments demonstrate a promising utility of our approach, accumulating 7.5$\times$ and 9.0$\times$ smaller tracking-errors than the two benchmark methods. Moreover, we assess the feasibility of our framework through implementation on physical robots in a mock wildfire monitoring scenario. The results of our experimental evaluations are presented in a supplementary video, available on \texttt{\url{https://youtu.be/zTR07cKlwRw}}.

\subsection{\textcolor{black}{General Applicability of the Proposed Framework}}
\label{subsubsec:ApplicabilityofProposedAlgorithms}
\noindent \textcolor{black}{While grounded in an application to wildfire monitoring, the proposed coordinated planning and collaborative field coverage frameworks for teams of autonomous robots are broadly applicable to other \textcolor{black}{domains and problems} in which one is attempting to perform intelligent tracking and monitoring of many targets with few robotic sensors in large-scale operations~\cite{freed2005intelligent}. Such applications include search-and-rescue~\cite{waharte2010supporting,rudol2008human,tomic2012toward}, tracking of wildlife poachers~\cite{olivares2015towards,gonzalez2016unmanned,bondi2018near}, oil spill surveillance in oceans~\cite{klemas2015coastal,marques2016oil,allen2008enhanced}, border patrol and protection~\cite{haddal2010homeland,freed2005intelligent} and even air traffic control for urban air mobility~\cite{vascik2018assessment,thipphavong2018urban,vascik2018scaling}. Here, we discuss these potential applications.}

\textcolor{black}{\textbf{Search and Rescue Missions--} Considering a scenario in which a natural disaster has struck a city, a team of UAVs can be used to keep monitoring various distant search sites for detecting survivors. Here, robots need to make sure to visit each search site frequently enough to not miss any signs of motion; here, our analytical temporal and measurement-uncertainty bounds for the case of static targets can provide a probabilistic guarantee for surveillance quality. This example can be extended to other search-and-rescue scenarios where humans (groups or individuals) are lost in forests or mountains and various sites need to be searched persistently or even for animal and wildlife monitoring and control, such as tracking specific animal species.}

\textcolor{black}{\textbf{Border Patrol and Protection--} Similar to the search-and-rescue example, border patrol and protection can be a relevant application for our framework in which aerial surveillance and tracking of multiple moving target (e.g., cars) can be performed by hovering \textcolor{black}{UAV}s. Various distant areas each including multiple targets can be monitored with a small team of UAVs which estimate the states (\textcolor{black}{i.e.}, position and velocity) of specific targets. For this application, our analytical bounds for dynamic targets can provide the minimum number of UAVs, required to ensure a high tracking quality of targets. A related example would be the tracking of wildlife poachers.}

\textcolor{black}{\textbf{Oil Spill Detection and Surveillance--} Detecting and surveilling oil spills in the oceans are also a closely relevant application for our frameworks. Similar to the aerial wildfire detection and monitoring, teams of autonomous UAVs can be deployed for large-scale oil spill surveillance, including multiple distant and dynamically growing spills, similar to the third case of our wildfire case-study, the moving-spreading fire areas.}

\section{Related Work}
\noindent Recent advances in UAV technology have opened up the possibility of providing real-time, high-quality fire information to firefighting teams across thousands of acres~\cite{afghah2019wildfire,de2020aerial,xing2019kalman}. 
However, coordinated control of UAV teams in this volatile setting provides particular challenges, such as decentralized controller design, task scheduling problems, large-scale communication, smoke detection and image stabilization, etc.~\cite{beard2006decentralized,mcintire2016iterated,nunes2015multi,choi2009consensus,allison2016airborne}.

Among the vision-based fire coverage methods~\cite{xing2019kalman, de2018real, chamoso2018computer, merino2012unmanned}, Merino \textit{et al.}~\cite{merino2010automatic} investigated the utility of visual and infrared cameras on UAVs to monitor the evolution of the firefront shape and then attempted to extend the proposed approach to a collaborative scheme performed by multiple UAVs. However, Yuan \textit{et al.}~\cite{yuan2015survey} note that vision-based approaches for fire monitoring and tracking struggle with image stabilization and camera obstruction, e.g., due to smoke, which produce significant errors in sensor measurements. Without addressing these errors properly, relying on such systems can be fatal.

\textcolor{black}{Our work is closely relevant to the cooperative multi-robot target tracking literature. A comprehensive review of the taxonomy and recent approaches on multi-robot target detection and tracking is provided by Robin and Lacroix in~\cite{robin2016multi}. An intensity function-based algorithm is presented in~\cite{jung2006cooperative} to generate a control law for tracking dynamic targets with a group of UAVs. The problem of dynamic distributed task allocation in ground-robot teams for coordinated target tracking is studied in~\cite{jin2019dynamic} through a bio-inspired approach. Mottaghi \textit{et al.}~\cite{mottaghi2007integrated} developed a particle filter-based approach to create a potential field to track a moving fire with multiple robots. Hausman \textit{et al.}~\cite{hausman2015cooperative} developed a probabilistic localization and control method for a UAV team with fixed number of robots which seeks to minimize the expected future uncertainty of the target position. Both mutual information and EKF's covariance are investigated as measures of uncertainty.}

\textcolor{black}{A review of the recent methods on the automation methods for wildfire remote sensing application via UAVs is presented in~\cite{bailon2020wildfire}, introducing and discussing three key metrics, i.e., situational awareness, decisional ability, and
collaboration ability, in the recent relevant literature.} Recent work by Sujit \textit{et al.}~\cite{sujit2007cooperative} proposed a cooperative approach to detect and monitor multiple spots of fire (i.e. hotspots) using two groups of detector and service UAV agents. Afghah \textit{et al.}~\cite{afghah2019wildfire}, proposed a distributed leader-follower coalition formation model to cluster a set of UAVs into multiple coalitions that collectively cover a designated area. Kumar \textit{et al.}~\cite{kumar2011cooperative} proposed a cooperative control algorithm to first cooperatively track the firefront shape for accurate situational awareness and then, autonomously fight the fire using fire suppressant fluid. Ghamry \textit{et al.}~\cite{ghamry2016cooperative} proposed a distributed algorithm with multiple stages such as search, confirmation and propagation monitoring for a team of UAVs to evenly distribute in a leader-follower manner to track an elliptical fire perimeter. Pham \textit{et al.}~\cite{pham2017distributed} designed a fire heat-intensity based distributed control framework for a team of UAVs to be capable of closely monitoring a wildfire in open space in order to track its development. Harikumar \textit{et al.}~\cite{harikumar2018multi} proposes a search and dynamic formation control framework for a multi-UAV system to efficiently search for a dynamic target in an unknown environment. 

\textcolor{black}{The majority of prior work on distributed control of UAVs, \textcolor{black}{e.g.} for MSN applications, focuses on maximizing the coverage of a particular area of interest~\cite{lee2015multirobot,lin2014optimal}, such as an area of wildfire~\cite{pham2018distributed,afghah2019wildfire}. Previous studies typically sought to maximize coverage (of a fire area) either through density-function based approaches~\cite{li2005distributed,zuo2019improved,santos2019decentralized,seraj2020coordinated} or by maximizing the average pixel density across a terrain~\cite{schwager2011eyes,pham2018distributed}. In the latter case, UAV-based wildfire coverage studies tend to adopt a fire-intensity function for their coverage problem formulation~\cite{pham2017distributed,lee2015multirobot,schwager2011eyes,seraj2020coordinated,li2005distributed,zuo2019improved,santos2019decentralized} to model areas of fire according to an intensity model.}

Many of the aforementioned approaches require an accurate fire-shape function~\cite{kumar2011cooperative} to work, or assume an enlarging elliptical perimeter for burning area to be monitored and tracked by moving UAVs~\cite{casbeer2006cooperative,sujit2007cooperative,ghamry2016cooperative,harikumar2018multi}. However, assuming a shape model for large-scale wildfires (when robot help is required) is not accurate~\cite{morvan2011physical}. Other approaches such as~\cite{pham2017distributed} use an ``interestingness'' function in order to enforce UAVs to look for areas of fire with predefined specifications. However, this function requires accurate, online measurements of heat intensity over the entire wildfire, which are often unavailable.

\textcolor{black}{Previous image pixel \textcolor{black}{density-based or intensity function-based} approaches to distributed control of UAVs for field coverage have typically sought to maximize the fire coverage, as their sole objective~\cite{pham2017distributed,afghah2019wildfire,lee2015multirobot,lin2014optimal}.} However, these approaches do not explicitly reason (i.e., through tracking and filtering) about a fire's state (i.e., position, velocity, scale and associated uncertainty), nor do they develop a predictive model for fire propagation by key parameters (e.g., fire-spread rate due to available fuel/vegetation and wind). Other approaches such as~\cite{de2020aerial} and~\cite{xing2019kalman} utilized Kalman filter to track fire position on input image data, using frames centroids as model inputs. Nevertheless, unlike these estimation-based approaches, we develop a predictive system with upper-bounds on the measurement-uncertainties that provides probabilistically-guaranteed performance.

Additionally, the fire-intensity based coverage approaches, such as~\cite{pham2017distributed} and~\cite{pham2018distributed}, tend to generate a density-function reflecting the heat-intensity in different parts of a terrain. Density-functions and Voronoi Tessellation based coverage approaches are amongst the most popular methods for static or dynamic field coverage in which robots are distributed and assigned to different parts of a map to maximize coverage~\cite{lee2015multirobot,schwager2011eyes,seraj2020coordinated,li2005distributed,zuo2019improved,santos2019decentralized}. Accordingly, fire-intensity based coverage approaches~\cite{pham2017distributed,pham2018distributed} typically tend to model the location of a firefront as the ``coolest'' or ``hottest'' part of a fire visible via infrared sensors~\cite{pham2017distributed} according to the wildfire scenario, but this assumptions are not always accurate~\cite{pastor2003mathematical} for the applications of continuous monitoring and tracking. \textcolor{black}{\textcolor{black}{These methods utilize a fire heat intensity model} to generate a time-varying density function and use artificial potential field to create driving force to control UAVs towards areas of minima (coolest part of fire) or maxima (hottest part of fire)~\cite{pham2017distributed}. \textcolor{black}{By using artificial potential fields,} such methods generate control inputs for each UAV at the low actuator level and thus, are different from our high-level coordinated planning for tracking moving targets. Intuitively, our high-level planner deals with the question of “what to do next?” while the mentioned low-level works deal with the problem of “how to get there?”.} Moreover, most of these methods are designed to cover as much of an area as possible with limited resources without systematically reasoning about the minimum required number robots for the task.

More closely along the line of our work, Bailon-Ruiz \textit{et al.}~\cite{bailon2018planning} proposed a model-based planning algorithms to monitor a propagating wildfire using a fleet of UAVs. The approach tailors a variable neighborhood search to plan surveillance trajectories for a fleet of fixed-wing \textcolor{black}{aircrafts} according to a \textcolor{black}{given} fire propagation model and a \textcolor{black}{given} wind model. The fire state predictions are performed for hours into the future by using the integrated models and are updated through \textcolor{black}{aircraft} observations. \textcolor{black}{Despite the similarities to our work, however, we did not consider the approach in~\cite{bailon2018planning} as a baseline for comparison, since the method assumes both a fire propagation model and a wind model to be given for making predictions and plans on the scale of hours into future and is designed for fixed-wing aircrafts with constrained motion dynamics.} In our approach, we enable UAVs to actively infer fire propagation model parameters through environment observations and plan accordingly, \textcolor{black}{whereas, in~\cite{bailon2018planning}, the state estimation is performed for extended temporal windows into the future. An active state estimation framework may be able to better cope with changes in fire dynamics over time.} \textcolor{black}{In addition, the method in~\cite{bailon2018planning} directly plans trajectories for fixed-wing UAVs with constrained differential-drive dynamics which are not applicable to our presumed omni-directional UAVs. The focus of our method, on the other hand, is to} provide a performance-guaranteed plan and \textcolor{black}{an upper-bound} on the number of UAV agents needed to monitor the fire areas without losing the track quality.

Furthermore, \textcolor{black}{data-driven and} learning-based approaches such as Reinforcement Learning (RL), have also been \textcolor{black}{broadly} applied to \textcolor{black}{the problem of dynamic field coverage} to enable collaborative monitoring of wildfires \cite{haksar2018distributed, xiao2020distributed, adepegba2016multi, seraj2020firecommander, zanol2019drone, seraj2022embodied}. \textcolor{black}{The problems of high-dimensional state-space and imperfect sensory information, which are common in the aerial wildfire monitoring application, are tackled in~\cite{julian2019distributed} and \cite{julian2018autonomous} for teams of fixed-wing aircrafts using two deep RL methods. A similar problem was investigated and tackled in~\cite{viseras2021wildfire} where a Q-learning~\cite{sutton2018reinforcement} was leveraged to learn one surveillance policy for a group of independent Q-learners. Ure \textit{et al.}~\cite{ure2015online} proposed a decentralized approach for the problem of multiple learning and collaborating agents in fire monitoring cases where agents estimate different models from their local observations, but they can share information by communicating model parameters. In~\cite{seraj2022learning, seraj2021heterogeneous}, a graph-based actor-critic~\cite{sutton2018reinforcement} method is introduced to learn efficient communication protocols for a group of cooperating heterogeneous agents (i.e., sensing and manipulating agents) performing wildfire fighting.}

While RL-based approaches occasionally show promising results in specific contexts, such as for small-scale fires~\cite{haksar2018distributed,julian2019distributed}, there are key limitations in terms of lack of formal guarantees on the boundedness of error, non-standard reward-function specification, scalability and inadaptability to domain shift, which are all limiting factors in application to safety-critical domains such as wildfire fighting.

\section{\textcolor{black}{Preliminaries: FARSITE Fire Propagation Mathematical Model}}
\label{subsec:simplifiedfarsite}
\noindent \textcolor{black}{Our coordinated multi-UAV planning framework relays on a dynamic target's approximate motion model to reason about the quality of service in robots' target-tracking mission. This approximate model is leveraged in Kalman filter to infer its latent parameters through environment observations. For our aerial wildfire coverage and firefront tracking case-study, we adopt the Fire Area Simulator (FARSITE) wildfire propagation mathematical model~\cite{finney1998farsite}.} 

\textcolor{black}{In this work, we leverage the simplified FARSITE model~\cite{pham2017distributed,seraj2020coordinated,pham2018distributed} to predict firefront propagation and use dynamic observations to infer latent parameters of the model. In our framework, UAVs quantify their estimation uncertainty by explicitly inserting the inferred parameters into the model and following a nonlinear uncertainty propagation law. The FARSITE wildfire propagation model used here can be replaced with any other parameterized model, such as the correctable fire simulation model introduced in~\cite{beachly2018fire}.}

Considering $ q_t^i $ as the location of $ i $-th firespot on firefront at time $ t $ and $ \dot{q}_{t}^i $ as a firefront's growth rate at location $q_t^i$ (i.e., fire propagation velocity), the wildfire propagation dynamics can be written as in Equation~\ref{eq:firemotionmodel}, where $ \delta t $ is the time-step and $ \dot{q}_{t}^i = \frac{d}{dt}\left(q_t^i\right) $ is a function of fire spread rate ($ R_t $, \textcolor{black}{i.e.}, fuel and vegetation coefficient), wind speed ($ U_t $), and wind azimuth ($ \theta_t $), which are available to our system through weather forecasting equipment.
\begin{equation}
    \label{eq:firemotionmodel}
    q_t^i = q_{t-1}^i + \dot{q}_{t-1}^i\delta t
\end{equation}In Equation~\ref{eq:firemotionmodel}, by ignoring the superscript $ i $ and without losing generality, the firefronts growth rate, $ \dot{q}_{t} $, can be estimated for each propagating spot on firefront by Equations~\ref{eq:qdot1}-\ref{eq:qdot2}, where $ \dot{q}_{t}^x $ and $ \dot{q}_{t}^x $ are first-order firefront dynamics for \textit{X} and \textit{Y} axes~\cite{finney1998farsite,seraj2020coordinated}.
\begin{align}
\dot{q}_{t}^x &= C(R_t, U_t)\sin(\theta_t) \label{eq:qdot1} \\
\dot{q}_{t}^y &= C(R_t, U_t)\cos(\theta_t)     \label{eq:qdot2}
\end{align}In above equations, $ C(R_t, U_t) $ is a firespot's velocity which is a function of the fuel/vegetation coefficient, $R_t$, and mid-flame wind speed, $U_t$. \textcolor{black}{We use equations from Finney~\cite{finney1998farsite} to calculate $C$ as in} Equation~\ref{eq:c}, in which \nobreak{\parskip0pt \small \noindent$ LB(U_t) = 0.936e^{0.256U_t} + 0.461e^{-0.154U_t} - 0.397 $} and \nobreak{\parskip0pt \small \noindent$ GB(U_t) = LB(U_t)^2 - 1 $}.
\begin{equation}
    \label{eq:c}
    C(R_t, U_t) = R_t\left(1-\frac{LB(U_t)}{LB(U_t) + \sqrt{GB(U_t)}}\right)
\end{equation}


\section{\textcolor{black}{Problem Formulation and Algorithmic Overview}}
\label{sec:ProbFprmulation&AlgOverview}
\subsection{Problem Statement}
\label{subsec:ProblemStatement}
\noindent \textcolor{black}{The proposed coordinated planning and collaborative coverage framework \textcolor{black}{with quality-of-service guarantees} is described and formulated in this \textcolor{black}{section} and is presented in Algorithm~\ref{alg:method}.} \textcolor{black}{We assume} that firefront loci are detected through vision or thermal cameras~\cite{merino2010automatic,yuan2015survey}. Moreover, \textcolor{black}{as discussed in Figures~\ref{fig:AreaCovergae} and \ref{fig:SafeHRI}, the UAV team is required to collectively monitor and track the fire (or generally moving targets) within several disjoint areas, such as areas of human-activity or human-defined areas of priority where high-quality information is indeed required. As such, we also assume that locations of these areas are known \textcolor{black}{a priori.}} 

\textcolor{black}{Upon receiving the request to monitor and track the fire in a total of $N_h$ areas of wildfire}, \textcolor{black}{an available \textcolor{black}{UAV}, $d$,} from a set of homogeneous UAVs, $ d\in\{\text{UAV}\}^{N_d} $ ($N_d$ is total number of \textcolor{black}{UAV}s), is required \textcolor{black}{to provide, online (or at within reasonable time frames) and high-quality information regarding the firefronts in the specified areas}. \textcolor{black}{The information here is referring to the estimated state information of a firespot, such as its location and velocity.} To this end, the UAV $d$  conducts a tour on firespots, $ \left(q_t^1, \cdots, q_t^{N_q}\right)\in\{Q_t\}^{N_q} $, where $ N_q $ is the total number of detected firespots \textcolor{black}{in all of $N_h$ areas.} This tour is performed to generate a graph, $ G_t $, on firespots and accordingly a path, $ \mathcal{P}_t^{g} $, on this graph for the UAV to follow as its tour-paths. \textcolor{black}{The graph, $G_t$, is created by considering the firespots, $q_t^{N_q}$, as its nodes and the straight path between them as its edges.} The path, $ \mathcal{P}_t^{g} $, must be created and updated such that the \textcolor{black}{UAV}, $d$, is capable of visiting \textcolor{black}{all $N_h$ areas} frequently enough \textcolor{black}{such that the estimated state information, i.e., firespot locations and velocities, would remain up-to-date between the UAV visits}. A list of inputs and outputs are summarized in Algorithm~\ref{alg:method}.

Since the velocity of the assigned UAV is limited to $v_{max}^d$ and fire's propagation parameters (\textcolor{black}{i.e.}, $R_t, U_t$ and $\theta_t$ in FARSITE model \textcolor{black}{in Equation~\ref{eq:firemotionmodel}}) \textcolor{black}{alter stochastically} over time, it will be challenging for a single \textcolor{black}{UAV} to monitor all firespots within \textcolor{black}{all $N_h$ areas} without losing the track quality, particularly if the monitoring areas are distant. As such, more UAVs need to be recruited and the monitoring tasks need to be divided among them. Note that, UAV resources are limited and thus, in large-scale operations such as wildfire fighting it is not feasible nor is efficient to simply use as many UAVs as possible to increase the quality \textcolor{black}{of the estimated information.} Accordingly, the number of required UAV agents need to be systematically bounded to efficiently monitor fire propagation within \textcolor{black}{all $N_h$ human-defined} areas of priority, while not losing the fire tracking quality. For this purpose, we propose leveraging online inference of wildfire dynamics and quantifying UAVs' accumulated estimation uncertainties about the fire model's parameters in order \textcolor{black}{to efficiently coordinate the UAV team.}
\textcolor{black}{For \textcolor{black}{the} readers' convenience, we provide Table~\ref{tab:notations} listing the key variables and notations used throughout the article.}

\begin{table}[!t]
\renewcommand{\arraystretch}{1.3}
\caption{\textcolor{black}{Summary of key nomenclature used \textcolor{black}{in our paper}.}}
\label{tab:notations}
\centering
\begin{tabularx}{\columnwidth}{lll}
\Xhline{2\arrayrulewidth}
 \textcolor{black}{\textbf{Notation}} & \textcolor{black}{\textbf{Domain}} & \textcolor{black}{\textbf{Definition and Properties}} \\
\Xhline{2\arrayrulewidth}
\textcolor{black}{$N_q$} & \textcolor{black}{$\mathbb{N}^+$} & \textcolor{black}{Total number of detected firespots (or ``targets'' in general)} \\
\hline
\textcolor{black}{$N_d$} & \textcolor{black}{$\mathbb{N}^+$} & \textcolor{black}{Total number of available \textcolor{black}{UAV}s (or ``robots'' in general)} \\
\hline
\textcolor{black}{$N_h$} & \textcolor{black}{$\mathbb{N}^+$} & \textcolor{black}{Total number of areas to be monitored and tracked} \\
\hline
\textcolor{black}{$\hat{}$} & \textcolor{black}{$-$} & \textcolor{black}{Accent used for estimated variables} \\
\Xhline{2\arrayrulewidth}
\textcolor{black}{$q_t^i$} & \textcolor{black}{$\mathbb{R}^{1\times2}$} & \textcolor{black}{Location of $i$-th firespot at time $t$ (\textcolor{black}{i.e.}, $q_t=\left[q_t^x, q_t^y\right]$)} \\
\hline
\textcolor{black}{$p_t^d$} & \textcolor{black}{$\mathbb{R}^{1\times3}$} & \textcolor{black}{Position of $d$-th \textcolor{black}{UAV} at time $t$ (\textcolor{black}{i.e.}, $p_t=\left[p_t^x, p_t^y, p_t^z\right]$)} \\
\hline
\textcolor{black}{$v_{max}^d$} & \textcolor{black}{$\mathbb{R}$} & \textcolor{black}{Maximum linear velocity of \textcolor{black}{UAV} $d$} \\
\Xhline{2\arrayrulewidth}
\textcolor{black}{$\mathcal{M}_t$} & \textcolor{black}{$\mathbb{R}$} & \textcolor{black}{Fire propagation model (or in general, a target's motion model)} \\
\hline
\textcolor{black}{$\mathcal{O}_t$} & \textcolor{black}{$\mathbb{R}$} & \textcolor{black}{\textcolor{black}{UAV}'s observation model (value at time $t$)} \\
\hline
\textcolor{black}{$T_{UB}$} & \textcolor{black}{$\mathbb{R}$} & \textcolor{black}{The upper-bound time required for a \textcolor{black}{UAV} to complete a tour in a designated area} \\
\hline
\textcolor{black}{$\text{URR}_t^q$} & \textcolor{black}{$\mathbb{R}$} & \textcolor{black}{The uncertainty residual ratio} \\
\hline
\textcolor{black}{$\text{S}_{t|t}$} & \textcolor{black}{$\mathbb{R}^{m\times m}$} & \textcolor{black}{The measurement residual covariance matrix; $m$ is the number of state variables} \\
\hline
\textcolor{black}{$\mathcal{A}_t^d$} & \textcolor{black}{$\mathbb{N}^{1\times 2}$} & \textcolor{black}{Assigned \textcolor{black}{UAV}-path pairs} \\
\Xhline{2\arrayrulewidth}
\end{tabularx}
\vspace*{-0.5cm}
\end{table}

\subsection{Algorithmic Overview}
\label{subsec:AlgorithmicOverview}
\begin{figure}[t!]
	\centering
	\includegraphics[width=\columnwidth]{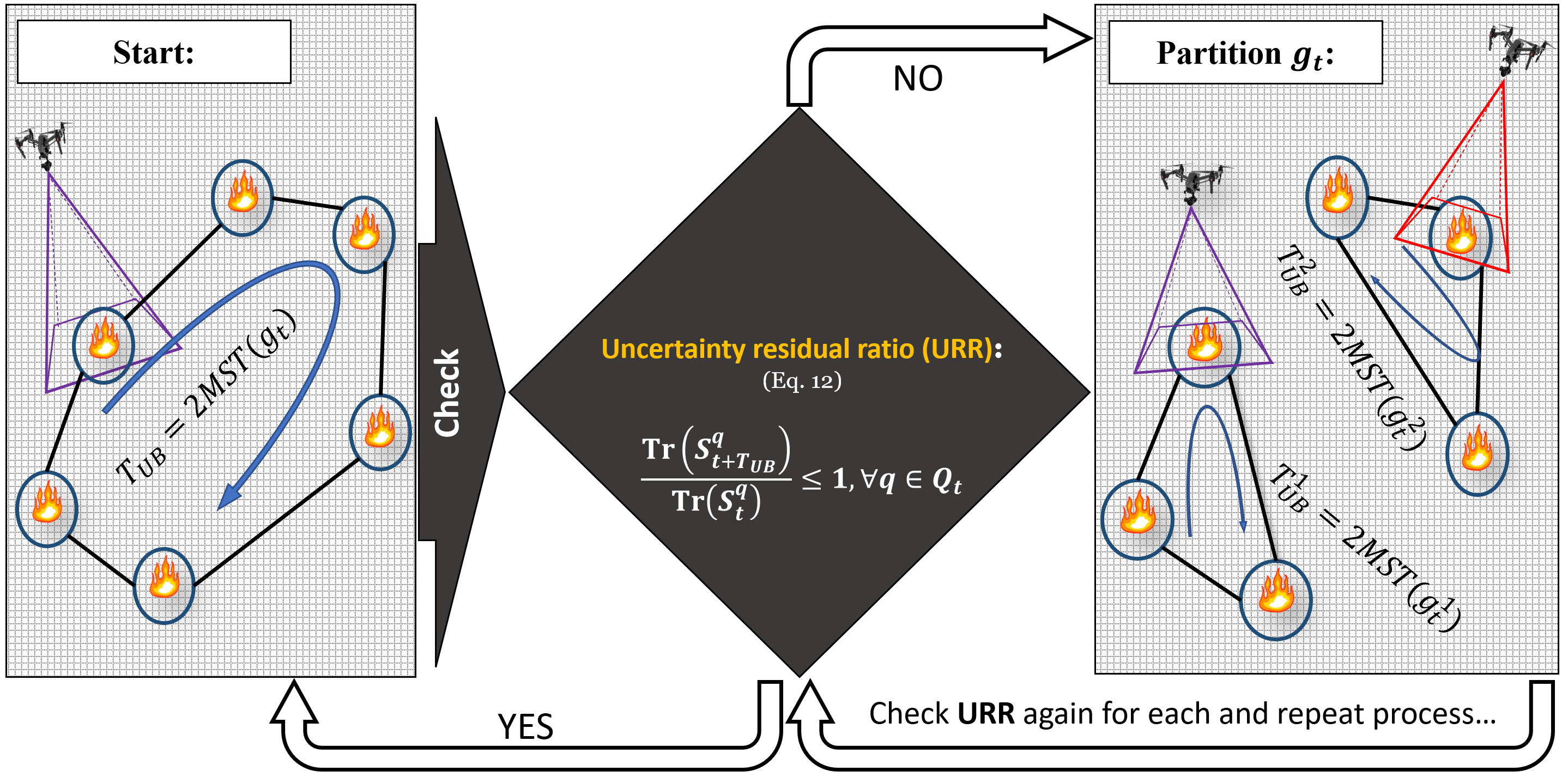}
	\caption{A flowchart diagram representing the URR checking process as the analytical safety condition. If the URR bound is not satisfied for a node in $g_t$, the graph is partitioned into two sub-graphs $g_t^1$ and $g_t^2$ \textcolor{black}{(i.e., through $k$-means clustering), and accordingly into two paths, and another \textcolor{black}{UAV} is called-in from the available UAVs in the team.} The process is repeated until URR is satisfied for all sub-graphs.}
	\label{fig:URRcheck}
	\vspace*{-0.5cm}
\end{figure}

\noindent Our solution to the aforementioned problem in Section~\ref{subsec:ProblemStatement} is overviewed here. Algorithm~\ref{alg:method} presents \textcolor{black}{our multi-UAV coordinated planning framework for active monitoring of dynamic environments, such as wildfire areas.} Figure~\ref{fig:URRcheck} represents a flowchart diagram of the core analytical process for \textcolor{black}{guaranteeing performance} in our proposed algorithm. 

Initially \textcolor{black}{areas that need to be monitored are} prioritized and divided from the rest of the map (Line 3). A selected UAV travels to these areas and gathers sensing information by flying over firespots and creating a graph, $ G_t $, with detected firespots as its vertices and a tour-path, $ \mathcal{P}_t^{g} $, by connecting the vertices on the generated graph as its edges (Lines 4-5). The graph $ G_t $ is an undirected graph generated by connecting each point to its closest neighboring point (\textcolor{black}{i.e.}, minimum spanning tree) and thus, the initial, quickly-generated graph, $ G_t $, and tour-path, $ \mathcal{P}_t^{g} $, are not efficient and therefore, need to be modified. In our approach, generating the revised graph $ G_t' $ and the path $ \mathcal{P}_t^{g'} $ on detected firespots within \textcolor{black}{specified areas} includes leveraging a Close-Enough Traveling Salesman Problem (CE-TSP) step to account for multiple fire-spots observable within UAVs' Field-of-View (FOV) (Line 6). \textcolor{black}{In CE-TSP, a UAV only \textcolor{black}{needs to get} ``close enough'' to fire-spots for state estimation rather than going to the exact location of each fire-spot.} Next, the UAV must determine through a \textit{feasibility test} that whether \textcolor{black}{monitoring all dynamic fire-spots within the $N_h$ areas can be guaranteed by taking the generated path, such that the track of none of the spots is lost.}

The performed feasibility test (Lines 9-13) is based on an analytical, probabilistic bound on the tracking error, which evaluates the measurement uncertainty residual about the state estimate, $ \hat{q}_t^i $, of a firespot, $ q_t^i $, at time $ t $, where $ i=1,\cdots,N_q $. The tracking-error residual bound which we refer to as $ \text{URR}_t^{\hat{q}} $, is evaluated through determining an upper-bound on the time, $T_{UB}$, required to cover each firespot in \textcolor{black}{the $N_h$ human-defined vicinities of priority (Line 7)}. In our feasibility test, $T_{UB}$ is compared to the maximum time allowable for each fire track to propagate before measurement, so that the post-measurement residual is less than the residual after the last time a given firespot was observed (Line 35). As shown in Figure~\ref{fig:URRcheck}, if the test is satisfied, the aforementioned process continues and a near-optimal tour is computed via a k-opt procedure. If the test fails, the UAV divvies up the responsibilities for covering the fire locations by partitioning graph $ G_t' $ into $ \langle g_1', g_2'\rangle $, recruiting another UAV to assist and repeating this process until $ \text{URR}_t^{\hat{q}} $ bound is satisfied for all sub-graphs of $ G_t' $. \textcolor{black}{The partitioning process can be performed through clustering approaches such as $k$-means.} UAVs' status are updated at the end for allocated/unallocated (i.e., 1/0 respectively) UAVs.

In the meantime, the unallocated team of UAVs can be utilized for \textcolor{black}{coordinated, distributed field coverage (\textcolor{black}{see Section~\ref{subsec:distributedcoverage} and Algorithm~\ref{alg:method1}}), for monitoring the rest of the wildfire area (i.e., areas that are not specified or prioritized). To this end, unallocated} UAVs generate a search graph and cluster the hotspots to generate a time-optimal path by leveraging the estimated temporal upper-bound $T_{UB}$.
\begin{algorithm}
	\footnotesize
	\SetKwData{NumUAV}{NumUAV}\SetKwData{TimeAvailable}{TimeAvailable}\SetKwData{MissionDuration}{MissionDuration}\SetKwData{True}{True}\SetKwData{False}{False}
	\SetKwFunction{CETSP}{CETSP}\SetKwFunction{UpdateDroneStats}{UpdateDroneStats}\SetKwFunction{UpdateMap}{UpdateMap}\SetKwFunction{CSP}{CSP}\SetKwFunction{Sense}{Sense}\SetKwFunction{ComputePath}{ComputePath}\SetKwFunction{FeasibilityTest}{FeasibilityTest}\SetKwFunction{Cluster}{Cluster}\SetKwFunction{TUB}{TUB}\SetKwFunction{CoordinatedCoverage}{CoordinatedCoverage}\SetKwFunction{len}{len}\SetKwFunction{MST}{MST}\SetKwFunction{kopt}{k-opt}
	\SetKwInOut{Input}{input}\SetKwInOut{Output}{output}
	\Input{Obtain the fire-map $ \{Q_t\}^{N_q} $ and total number of \textcolor{black}{areas to cover} $N_h$, list of all \textcolor{black}{UAV}s $ \{\text{UAV}\}^{N_d} $, set of UAV poses, velocities and observation model $\langle \{p_t\}^d, v_{max}^d, \mathcal{O}_t\rangle$, fire propagation model $ \mathcal{M}_t $}
	\Output{Assigned \textcolor{black}{UAV}-path pairs $ \{\mathcal{A}_t^d\} $}
	\BlankLine
	
	// main loop //
	
	\While{MissionDuration}{
		
		Update fire-map and priority areas: $ \langle\{Q_t\}^{N_q^1}, \{Q_t\}^{N_q^0}\rangle \leftarrow $ \UpdateMap{$\{Q_t\}^{N_q}$}
		
		Estimate fire states and dynamics: $ \{\hat{q}_t\} \leftarrow $ \Sense{$\{Q_t\}^{N_q}, \{p_t\}^d, \mathcal{M}_{t}, \mathcal{O}_{t}$}
		
		Form a graph and compute path: $ \langle\text{G}_t, \mathcal{P}_t^{g}\rangle \leftarrow $ \ComputePath{$\{\hat{q}_t\}$}
		
		Modify graph of estimated spots: $ \langle\text{G}_t', \mathcal{P}_t^{g'}\rangle \leftarrow $ \CETSP{$\text{G}_t, \mathcal{P}_t^{g}$}
		
		Compute $ T_{UB} $ for $ d\in\{\text{UAV}\}^d $ to cover $ \text{G}_t' $: $ T_{UB}^{g'} \leftarrow $ \TUB{$\mathcal{P}_t^{g'}, \{\hat{q}_t\}, v_{max}^d$}
		
		Create a set from graphs and their corresponding $ T_{UB} $s: $ \{\mathcal{G}\} \leftarrow \langle\text{G}_t', T_{UB}^{g'}\rangle$
		
		\While{$ \exists g \in \{\mathcal{G}\} ~\text{s.t.} \False \leftarrow $ \FeasibilityTest{$\{\mathcal{G}\}, \{\hat{q}_t\}$}}{
			
			$ \langle\{g'_1, g'_2\}, \{\mathcal{P}_{g_1}', \mathcal{P}_{g_2}'\}\rangle \leftarrow $ \CETSP{\Cluster{$\{\mathcal{G}\}^g, 2$}}
			
			$ \{T_{UB}^{g_1'}, T_{UB}^{g_2'}\} \leftarrow $ \TUB{$\{\mathcal{P}_{g_1}', \mathcal{P}_{g_2}'\}, \{\hat{q}_t^g\}\in g, v_{max}^d$}
			
			$ \{\mathcal{G}\} \leftarrow \langle\{g'_1, g'_2\}, \{T_{UB}^{g_1'}, T_{UB}^{g_2'}\}\rangle $
			
		}
		
		Assign \textcolor{black}{UAV}s to paths: $ \{\mathcal{A}_t^d\} = \langle\mathcal{P}_t^{g'}, d\rangle, \forall d \in \{\text{UAV}\}^{N_d} $

	}
	// inner functions //
	
	\textbf{def} ~\Sense{$\{Q_t\}^{N_q}, \{p_t\}^d, \mathcal{M}_{t}, \mathcal{O}_{t}$}\textbf{:}~~~// dynamic fire state estimations
	
	\hspace{0.5cm}$\hat{q}_{t+1} = \argmax_{q_{t+1}} \rho\left(q_{t|t-1}, p_{t|t-1}, \mathcal{M}_{t}, \mathcal{O}_{t}\right)$
	
	\textbf{def} ~\ComputePath{$\{\hat{q}_t\}$}\textbf{:}~~~// generate and optimize path
	
	\hspace{0.5cm}$ \text{G}_t, \mathcal{P}_t^{g} \leftarrow $ \MST{$\{\hat{q}_t\}$}~~~// \MST{.}: minimum spanning tree graph
	
	\hspace{0.5cm}\textbf{while} \TimeAvailable \textbf{do}
	
	\hspace{0.7cm}$\vert$\hspace{0.15cm}$ \mathcal{P}_t^{g} \leftarrow $ \kopt{$\text{G}_t, \mathcal{P}_t^{g}$}~~~// \kopt{.}: to optimize path
	
	\hspace{0.5cm}\textbf{end}
	
	\textbf{def} ~\TUB{$\mathcal{P}_t^g, \{\hat{q}_t\}, v_{max}^d$}\textbf{:}~~~// temporal service upper-bound
	
	\hspace{0.5cm}Determine fire scenario according to estimated states $ \hat{q}_t^i $
	
	\hspace{0.5cm}\textbf{switch}
	
	\hspace{0.8cm}\textbf{Case (1):} $ T_{UB}^{C1} \leftarrow \frac{2\len(\mathcal{P}_t^g)}{v_{max}^d} $ // See Equation~\ref{eq:T_UBcase1} for details
	
	\hspace{0.8cm}\textbf{Case (2):} $ T_{UB}^{C2} \leftarrow \frac{8\zeta^\alpha(|\text{G}_t|-1)\len(\mathcal{P}_t^g)}{v_{max}^d\left(1-4\zeta^{\alpha}\left(|\text{G}_t|-1\right)\right)} $ // See Equation~\ref{eq:T_UBcase2} for details
	
	\hspace{0.8cm}\textbf{Case (3):} $ T_{UB}^{C3} \leftarrow \frac{-\beta+\sqrt{\beta^2-4\gamma\delta}}{2\gamma} $ // See Equation~\ref{eq:T_UBCase3} for details
	
	\textbf{def} ~\FeasibilityTest{$\{\mathcal{G}\}, \{\hat{q}_t\}$}\textbf{:}
	
	\hspace{0.5cm}\textbf{if} $\frac{\Tr\left(S_{t+T_{UB}|t}\right)}{\Tr\left(S_{t|t-1}\right)} \leq 1, \forall g \in \{\mathcal{G}\}$, \textbf{return} \True \textbf{else, return} \False

	\caption{\textcolor{black}{Stages of the proposed multi-UAV coordinated dynamic field monitoring framework (pre-specified areas of priority) with guaranteed performance}.}
	
	\label{alg:method}
	
\end{algorithm}

\section{\textcolor{black}{Coordinated Planning for Monitoring with Guaranteed Quality-of-Service}}
\label{sec:method}
\noindent Our objective is to propose a \textcolor{black}{performance}-guaranteed \textcolor{black}{coordinated planning framework for collaborative wildfire tracking}. Particularly, our approach consists of a probabilistically-\textcolor{black}{guaranteed}, realtime coordination algorithm that provides human firefighters \textcolor{black}{with real-time information about fire states within the specified areas}. We then develop a predictive distributed coverage method for large-scale dynamic field coverage based on our probabilistic bounds to improve the performance in both wildfire area coverage and firefront tracking.

\subsection{Online Inference of Wildfire Dynamics}
\label{subsec:errorpropagation}
\noindent To perform a real-time inference of wildfire dynamics, we utilize \textcolor{black}{the Adaptive Extended Kalman Filter (AEKF)} to predict a distribution over the states of observed firespots and accordingly, a measurement covariance for each \textcolor{black}{firespot (i.e., each discretized point of firefront)} through non-linear error propagation~\cite{kalman1961new,akhlaghi2017adaptive}. \textcolor{black}{In a linear system with Gaussian noise, the Kalman filter is optimal~\cite{simon2006optimal}. In a system that is nonlinear, the Kalman filter can be used for state estimation, but other methods such as the particle filter may give better results, normally at the price of significant additional computational effort~\cite{simon2006optimal}. In our work, we \textcolor{black}{develop} computationally lightweight algorithms to scale to large numbers of UAVs and coverage areas. As such, we leverage a Kalman filter as it provides such low-complexity algorithm while also performing satisfactorily.}

Considering $ q_{t}=\left[q_t^x, q_t^y\right] $ as the location of \textcolor{black}{firefronts on the ground at time, $t$,} and $ p_{t}=\left[p_t^x, p_t^y, p_t^z\right] $ as the UAV pose, we seek to estimate a distribution over a firespot's location one step forward in time, $ \hat{q}_{t+1} $, given the current firefront distribution and the previous measurement (shown as $ q_{t|t-1} $ in our notations throughout the paper), fire propagation model \textcolor{black}{with parameters at time, $t$,} $ \mathcal{M}_{t} $, and UAV observation model of the field, $ \mathcal{O}_{t} $. Considering the parameters in fire propagation model (\textcolor{black}{i.e.}, $R_t, U_t$ and $\theta_t$ in FARSITE model \textcolor{black}{in Equation~\ref{eq:firemotionmodel}}), the firespot state estimation problem can be formulated as maximizing the joint probability density function (PDF), $\rho$, in Equation~\ref{eq:jointPDF}. \textcolor{black}{Equation~\ref{eq:jointPDF} \textcolor{black}{describes} a joint probability density function for estimating the next-step position of a fire-spot, given the current firefront distribution and the previous measurement, fire propagation model, and UAV observation model. Therefore, maximizing this PDF means finding a value of the random variable (the value estimate) that can occur with the highest probability.}
\begin{equation}
\label{eq:jointPDF}
\begin{aligned}
\hat{q}_{t+1} = \argmax_{q_{t+1}} \rho\left(q_{t|t-1}, p_{t|t-1}, \mathcal{M}_{t}\left(R_{t|t-1}, U_{t|t-1}, \theta_{t|t-1}\right), \mathcal{O}_{t}\left(q_{t|t-1}, p_{t|t-1}\right)\right)
\end{aligned}
\end{equation}
\begin{figure}[t!]
	\centering
	\includegraphics[width=0.65\columnwidth]{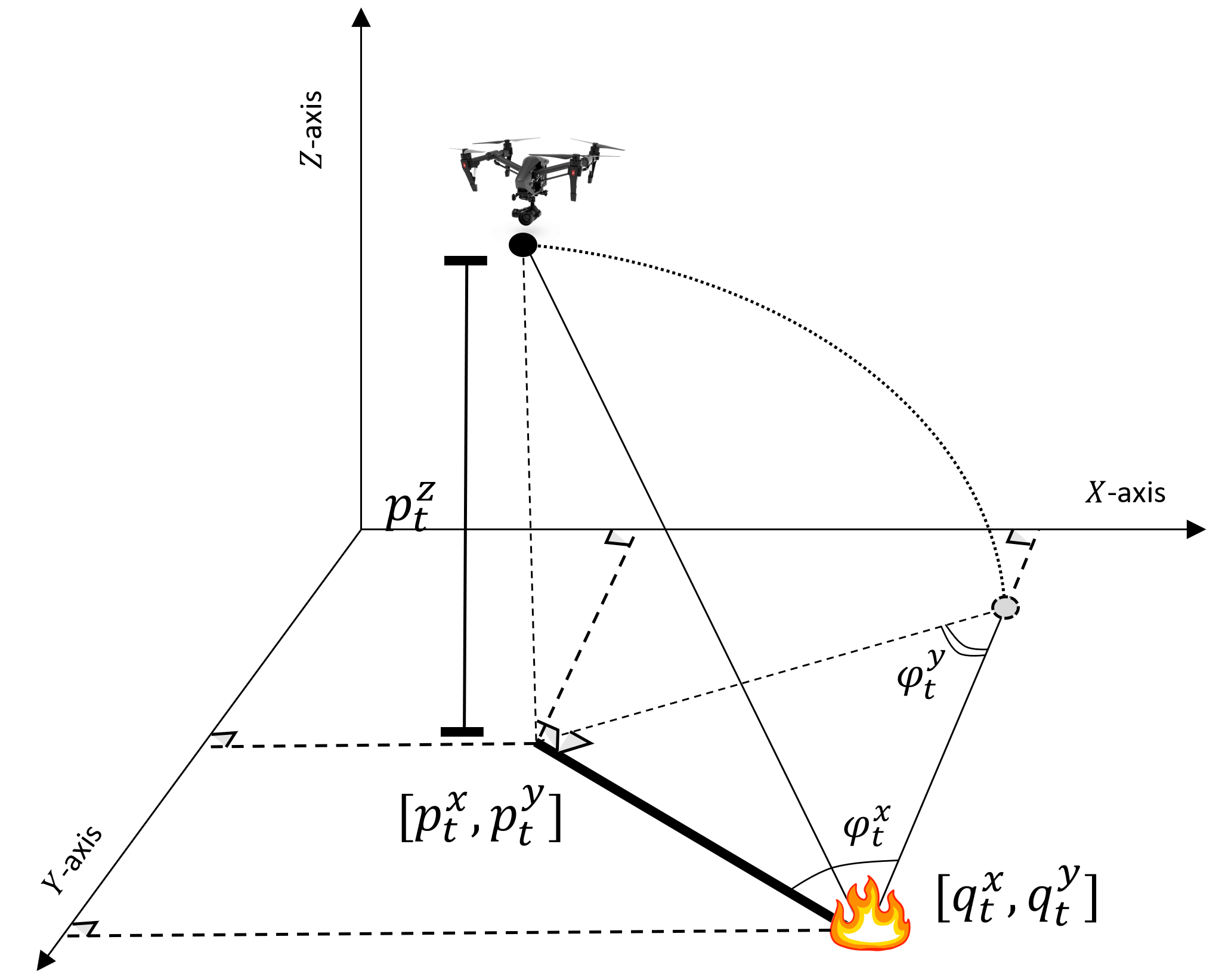}
	\caption{The observation model of a hovering UAV with position $ p_{t}=\left[p_t^x, p_t^y, p_t^z\right] $, with respect to a target (e.g., firespot) on the ground with location $ q_{t}=\left[q_t^x, q_t^y\right] $. \textcolor{black}{The angle parameter, $\varphi_t$, represents the angle by which the UAV \textcolor{black}{observes} the target on the ground. We note that, although the angle-parameters, $\varphi_t^y$ and $\varphi_t^x$, are complementary angles, we utilize them both since \textcolor{black}{we ultimately leverage} these angles for tracking 2D pose uncertainty \textcolor{black}{in $\langle x, y \rangle$} (and not just angle uncertainties).}}
	\label{fig:ObsMdl}
	\vspace*{-0.5cm}
\end{figure}

The PDF in Equation~\ref{eq:jointPDF} is calculated through the AEKF estimator. The process model, $ \mathcal{M}_{t} $, predicts states given its parameters (\textcolor{black}{i.e.}, $ R_{t}, U_{t}, \theta_{t} $) by leveraging the fire propagation model (Section~\ref{subsec:simplifiedfarsite}). The observation model, $ \mathcal{O}_{t} $, maps the predicted state estimates to observation space.\textcolor{black}{As demonstrated in Figure~\ref{fig:ObsMdl}, considering a dynamic target on the ground, $ q_t=[q_t^x, q_t^y] $, moving according to a nonlinear model, $ \mathcal{M}_t $ (\textcolor{black}{i.e.}, Eq.~\ref{eq:firemotionmodel}), at each time, $ t $, the observation mapping along X and Y axes, $ \mathcal{O}_t^x $ and $ \mathcal{O}_t^y $, of a flying perception agent with pose, $ p_t=[p_t^x, p_t^y] $ and altitude $p_t^z$, with respect to the target's location can be shown as in Equations~\ref{eq:obsmdl_x}-\ref{eq:obsmdl_y}.
\begin{align}
   \label{eq:obsmdl_x}
   \mathcal{O}_t^x: \{q_t, p_t\}\rightarrow & \varphi_t^x: \left\Vert q_t^z-p_t^z\right\Vert_2 = \left\Vert q_t-p_t\right\Vert_2\tan\varphi_t^x \\
   \mathcal{O}_t^y: \{q_t, p_t\}\rightarrow & \varphi_t^y: \left\Vert q_t-p_t\right\Vert_2 = \left\Vert q_t^z-p_t^z\right\Vert_2\tan\varphi_t^y
   \label{eq:obsmdl_y}
\end{align}}Moreover, we incorporate process and observation noises to account for error induced by methodological uncertainties (\textcolor{black}{i.e.}, fire propagation model and UAV observation model inaccuracies) and other sensor errors (\textcolor{black}{i.e.}, camera and weather forecasting equipment), which are modeled as white noise with covariances $ \Lambda_t $ and $ \Gamma_t $, respectively. We leverage an adaptive update framework~\cite{akhlaghi2017adaptive} for process and observation noise covariances in standard EKF equations, which introduces \textit{innovation}- and \textit{residual}-based updates for $ \Lambda_t $ and $ \Gamma_t $, respectively. \textcolor{black}{The adaptive updates, as introduced in~\cite{akhlaghi2017adaptive}, and shown in Equations~\ref{eq:adaptiveQ}-\ref{eq:adaptiveR}, relax the assumption of constant \textcolor{black}{process and observation} covariances, $ \Lambda_t $ and $ \Gamma_t $, and reduce sensitivity to their initialization \textcolor{black}{by adaptively estimating these parameters based on innovation and residual to improve the dynamic state estimation accuracy of the standard EKF.}}
\textcolor{black}{
\begin{align}
\Lambda_t &= \gamma \Lambda_{t-1} + \left(1-\gamma\right)\left(K_t\tilde{d}_t\tilde{d}_t^TK_t^T\right) \label{eq:adaptiveQ} \\
\Gamma_t &= \gamma \Gamma_{t-1} + \left(1-\gamma\right)\left(\tilde{y}_t\tilde{y}_t^T + H_tP_{t|t-1}H_t^T\right) \label{eq:adaptiveR}
\end{align}}\textcolor{black}{In above equations, $\gamma$ is the update step-size, $ \tilde{y}_t $ is the measurement residual, and $\tilde{d}_t$ is the measurement innovation. \textcolor{black}{The measurement residual is defined as the difference between actual measurement and the predicted measurement using the posterior, while the measurement innovation is defined as the difference between the actual measurement and its predicted value.} Moreover, $ K_t = P_{t|t-1}H_t^TS_t^{-1} $ is the near-optimal Kalman gain, in which, $ P_{t|t-1} $ is the predicted covariance estimate, $ H_t $ is the observation Jacobian matrix and $ S_t $ is the covariance residual. Through Equations~\ref{eq:adaptiveQ}-\ref{eq:adaptiveR}, $ \Lambda_t $ and $ \Gamma_t $ are adaptively estimated as the Kalman filter leverages its observations to improve the predicted covariance matrix, $ P_{t|t-1} $.} Accordingly, to derive the AEKF uncertainty (\textcolor{black}{i.e.}, measurement-residual) propagation equations, we define the process state vector as $ \Vec{\Theta}_t = \left[ q_t^x, q_t^y, p_t^x, p_t^y, p_t^z, R_t, U_t, \theta_t \right]^T $ and $ \Vec{\Phi}_t = \left[\varphi_t^x, \varphi_t^y, \hat{R}_t, \hat{U}_t, \hat{\theta}_t\right]^T $ as the mapping vector\footnote{Note that the state variables in $\Vec{\Theta}_t$ and $\Vec{\Phi}_t$ include firespot and UAV positions, $q_t$ and $p_t$, as well as $R_t, U_t$ and $\theta_t$ which are the FARSITE wildfire propagation model parameters. While these variables are generally application-dependent, we emphasis that $R_t, U_t$ and $\theta_t$ are specific to FARSITE (see Section~\ref{subsec:simplifiedfarsite}) model and can be replaced with other measurable model parameters, in case FARSITE is replaced with other fire propagation models such as the correctable fire simulation model in~\cite{beachly2018fire}.} through which the state estimates are translated into a unified \textit{angle}-parameters, as firespots on the ground are perceived by hovering UAVs (Figure~\ref{fig:ObsMdl})~\cite{seraj2020coordinated}. Eventually, the model and observation measurement uncertainties propagated by AEKF estimation can be shown as in Equations~\ref{eq:ekfModelUncertainty1}-\ref{eq:ekfObservationUncertainty1}.
\begin{align}
P_{t|t-1} &= F_tP_{t-1|t-1}F_t^T + \Lambda_t \label{eq:ekfModelUncertainty1} \\
\text{S}_{t|t} &= H_tP_{t|t-1}H_t^T + \Gamma_t \label{eq:ekfObservationUncertainty1}
\end{align}Here, $ P_{t|t-1} $ is the predicted covariance estimate and $ \text{S}_{t|t} $ is the innovation (or residual) covariance matrix. Moreover, $ F_t $ and $ H_t $ are the process and observation Jacobian matrices which include the gradients of FARSITE model equations in Equations~\ref{eq:qdot1}-\ref{eq:qdot2} with respect to all state variables in $ \Vec{\Theta}_t $ and the gradients of UAV's observation model in \textcolor{black}{Equations~\ref{eq:obsmdl_x}-\ref{eq:obsmdl_y}} with respect to all observation space variables in $ \Vec{\Phi}_t $, respectively (see~\cite{seraj2020coordinated} for derivations).

\subsection{\textcolor{black}{Guaranteeing the Quality of Service}}
\label{subsec:humansafetymodule}
\noindent Here, we first present the CE-TSP solution for generating more efficient tour-paths for UAV agents. Next, we introduce and elaborate on our analytical \textcolor{black}{condition for performance guarantee}, \textcolor{black}{i.e.}, the Uncertainty Residual Ratio (URR) bound, and then, we derive the analytical temporal upper-bounds for a probabilistically-guaranteed coordination based on URR.

\subsubsection{Close-enough Traveling Salesman Problem \textcolor{black}{(CE-TSP)}}
\label{subsubsec:CETSP}
\textcolor{black}{When a UAV agent is tasked to monitor (\textcolor{black}{i.e.}, estimate the firefront states and closely track each firespot), the first step in our coordinated planning module is to generate a search graph, $G_t$, with firefront points within the \textcolor{black}{specified areas} as the vertices and distances in between as the edges. To this end,} we leverage the Close-Enough Traveling Salesman Problem (CE-TSP) with Steiner zone~\cite{wang2019steiner} variable neighborhood search where the agent only gets ``close enough'' to each fire-point instead of visiting their exact locations\footnote{Note that, this step might not be required in applications other than aerial wildfire monitoring, in which instead of $N_h$ fire areas, $N_h$ specific moving points/targets need to be monitored. Accordingly, the CE-TSP step in our framework can be replaced with a regular TSP~\cite{applegate2006traveling}.}.

\textcolor{black}{In the application of aerial wildfire propagation monitoring, at each time-step, a UAV can observe multiple \textcolor{black}{firespots (\textcolor{black}{i.e.}, nodes)} within its \textcolor{black}{FOV} simultaneously. As such, it is not required for the UAV to} travel to the exact location of each firefront point. Instead, \textcolor{black}{the UAV} first identifies the overlapping $\Delta$-disks between $ k $ firespots as their corresponding Steiner zone~\cite{wang2019steiner} and \textcolor{black}{then chooses} the centroid node from the identified Steiner areas as a new vertex in \textcolor{black}{its search graph} instead of the \textcolor{black}{original} $ k $ points. Through this process the original graph $G_t$ is modified into a revised graph $G_t'$ (Line 5 in Algorithm~\ref{alg:method}). \textcolor{black}{The variable $k$ can be tuned empirically and based on the background application.}


\subsubsection{Analytical \textcolor{black}{Performance Guarantee} Condition: Uncertainty Residual Ratio (URR)}
\label{subsubsec:URR}
\noindent \textcolor{black}{When the modified search graph, $G_t'$, is generated, the UAV performs tours on this graph by visiting each graph-node.} Upon reaching a new node, the UAV updates the firespot's state estimate and calculates two quantities: (1) the analytical upper-bound time, $T_{UB}$, required to complete \textcolor{black}{a tour on $G_t'$ and arrive back at the current node considering the new state estimates} (described in Section~\ref{subsec:T_UB}) and (2) the current measurement residual covariance through Equation~\ref{eq:ekfObservationUncertainty1}. Leveraging these two parameters, the UAV propagates the uncertainty residual \textcolor{black}{for} $ T_{UB} $ steps into the future by repeatedly applying AEKF's \textcolor{black}{prediction step (i.e., Equation~\ref{eq:ekfModelUncertainty1}) $ T_{UB} $ times, and then performing an update step at the end (i.e., Equation~\ref{eq:ekfObservationUncertainty1}). In this way, we emulate gathering a measurement again and compute the innovation covariance at the end of the prediction steps.} We introduce the Uncertainty Residual Ratio ($ \text{URR}_t^{\hat{q}} $) in Equation~\ref{eq:URR1}, in which $\Tr(.)$ represents the trace operation.
\begin{equation}
\text{URR}_t^{\hat{q}} = \frac{\Tr\left(S_{t+T_{UB}|t}\right)}{\Tr\left(S_{t|t-1}\right)}~~~ \text{and}~~~ \text{URR}_t^{\hat{q}} \leq 1, \forall q \in \{Q_t\}^{N_q} \label{eq:URR1}
\end{equation} 

The $ \text{URR}_t^{\hat{q}} $ \textcolor{black}{bound} is an indicator of the scale to which the UAV agent is capable of tracking the \textcolor{black}{firespot, $ q_t $, without} losing tracking information, \textcolor{black}{while performing tours on the search graph, $G_t'$}. \textcolor{black}{An unsatisfied URR bound (\textcolor{black}{i.e.}, a URR \textcolor{black}{value} greater than one)} indicates growing uncertainty and demonstrates a quickly growing (or propagating) wildfire, about which the UAV will not be capable of providing \textcolor{black}{online state-estimates and without loosing any track information,} while completing a tour \textcolor{black}{on $G_t'$}. Similarly, a URR smaller than one for all firespots in the current graph indicates that online information can be provided \textcolor{black}{by the UAV while keeping track of all propagating firespots. We reiterate that, the goal of the UAV team here is to provide firefighters with the online, high-quality information about the propagating firefront within the specified areas such that these information can be used in real-time for strategizing firefighting plans.}

\textcolor{black}{\textcolor{black}{To maintain control over the measurement uncertainty, we posit that the UAV observers would want the measurement uncertainty residual} with respect to a target on the ground \textcolor{black}{not to} increase \textcolor{black}{from} $t=t_0$ \textcolor{black}{to} \textcolor{black}{$t=t_0 + kT_{UB}$ for any positive integer constant $k$} if the UAV observes the target from the same relative position. The reason is that the measurement uncertainty residual as computed by the EKF in our formulation is only dependent on the relative distance between the observer and the target and is independent of time (please refer to the provided Appendix for a mathematical proof and a detailed discussion). As such, the bound in URR ratio determines if the current uncertainty residual (the denominator \textcolor{black}{of Eq.~\ref{eq:URR1}}) is greater or smaller than the uncertainty residual at time $t+t_{UB}$ (the numerator \textcolor{black}{of Eq.~\ref{eq:URR1}}) when the UAV completes the tour and revisits the current target. Accordingly, if the ratio is greater than one, it means that the uncertainty residual is increasing, indicating the UAV is falling behind on monitoring the target and the tracking quality is getting worse.}

As demonstrated in flowchart diagram in Figure~\ref{fig:URRcheck}, if the URR bound \textcolor{black}{is not satisfied for a graph node ($\text{URR}_t^q > 1~~\text{for}~~ q\in\{Q\}^{N_q}$),} the generated \textcolor{black}{search graph, $ G_t' $,} is partitioned into two \textcolor{black}{smaller sub-graphs $ \langle g_1', g_2'\rangle $ (i.e., through $k$-means clustering), and accordingly two paths $ \langle \mathcal{P}_{g_1}', \mathcal{P}_{g_2}'\rangle $.} \textcolor{black}{Therefore, the UAV divvies up the monitoring task into smaller portions and then recruits another \textcolor{black}{UAV} from the available UAVs in the team to collaborate by taking over one of the sub-graphs for monitoring.} This process is repeated for all sub-graphs and all of their nodes until $ \text{URR}_t^{\hat{q}} $ is satisfied for all $ q\in\{Q_t\}^{N_q} $ and $ t $. \textcolor{black}{We note that, in our formulation, the UAVs only begin the surveillance and tracking the URR bound when they arrive to the determined sub-graph. As such, to minimize the time it takes a UAV to arrive to its assigned sub-graph we solve \textcolor{black}{an optimization problem}. Each \textcolor{black}{UAV} is assigned to one partition by solving a \textcolor{black}{minimization problem} with \textcolor{black}{UAV}s as variables, partitions as domains, and distance to the partition centroid as constraints. This way, the closest available UAV is selected for a sub-graph. Additionally, as described in Section~\ref{subsubsec:CETSP}, in cases where several firespots can be observed in a UAV's FOV at once, the URR bound is computed for the centroid node from the identified Steiner areas.}

The introduced URR bound in Equation~\ref{eq:URR1} depends on the upper-bound traverse time, $ T_{UB} $, which itself is dependent on \textcolor{black}{two major factors: (1) the maximum linear velocity of the UAV ($v_{max}^d$) and (2) the propagation rate} of the fire (Or in general, the motion velocity of any dynamic target which is subject to monitoring). \textcolor{black}{As such, the relations between $ T_{UB} $ and the aforementioned two factors need to \textcolor{black}{be} derived analytically for all possible scenarios. The following \textcolor{black}{section} is dedicated to \textcolor{black}{analyzing} such scenarios in the face of the aerial fire monitoring domain and the respective analytical temporal bounds are presented.}

\subsubsection{\textcolor{black}{Probabilistic Temporal Upper-Bound for Service ($T_{UB}$)}}
\label{subsec:T_UB}
\begin{figure}[t!]
	\centering
	\includegraphics[height=6cm, width=\columnwidth]{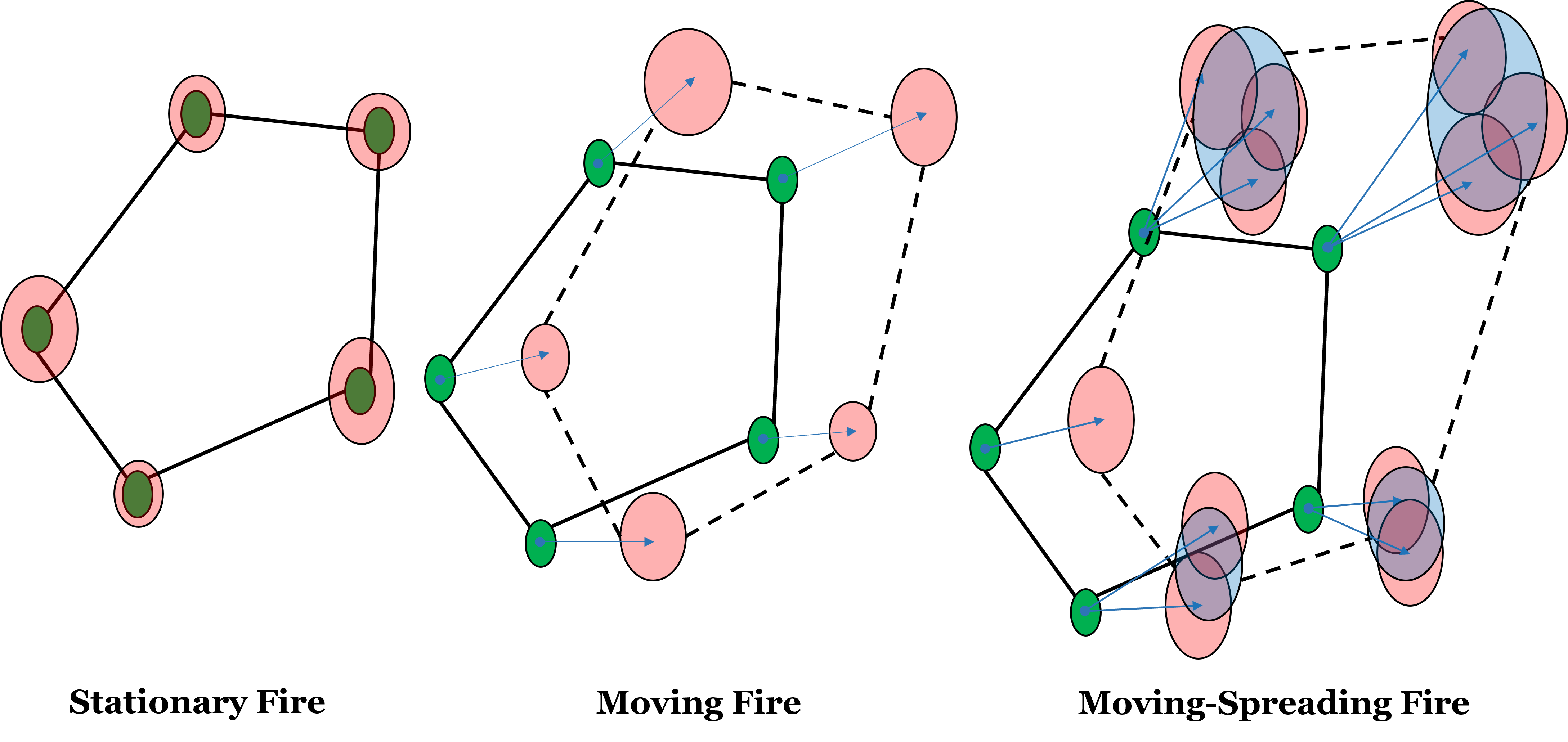}
	\caption{\textcolor{black}{The three considered scenarios in which, a fire (i.e., target) could be stationary (left), moving (middle) or \textcolor{black}{moving-and-spreading} (right).} \textcolor{black}{Green dots show a firespot's current location and red circles represent the distribution over the firespot's next-step location.} Blue circles show the Steiner zones. Note that the black, solid and dashed lines represent a UAV's current and next-step tour paths\textcolor{black}{, respectively.}}
	\label{fig:WildfireScenarios}
	\vspace*{-0.5cm}
\end{figure}

\noindent In this section, we \textcolor{black}{derive} a probabilistic upper-bound, $T_{UB}$, on the time required by a \textcolor{black}{UAV} to service (i.e., \textcolor{black}{visit once and estimate states}) each fire location. \textcolor{black}{The $T_{UB}$} is used in Equation~\ref{eq:URR1} to determine whether a \textcolor{black}{UAV} can be probabilistically guaranteed to service each fire location fast enough to ensure a bounded track residual for each \textcolor{black}{firespot}. \textcolor{black}{As \textcolor{black}{described} in Section~\ref{subsubsec:URR}, the $ T_{UB} $ is dependent on two major factors: (1) the maximum linear velocity of the UAV ($v_{max}^d$) and (2) the propagation velocity and growth rate of firespots.} \textcolor{black}{As such,} we derive three bounds, one for each of the following \textcolor{black}{possible} scenarios: (1) stationary target points\footnote{The scenario designs are motivated such that they expand the applicability of our framework to domains other than wildfire \textcolor{black}{monitoring}, and thus, here we use the term \textit{target points} instead of \textit{firespots}.}, (2) moving target points, and (3) moving-spreading target points. Figure~\ref{fig:WildfireScenarios} depicts \textcolor{black}{the three mentioned scenarios subject to our study}.

For the derivations, we reason about the velocity of firespots (e.g., target points)\textcolor{black}{, $\zeta$,} at the $\alpha$ confidence level\textcolor{black}{, such that, $ Pr[\zeta < \zeta^{\alpha}] = 1 - \alpha $ where smaller $ \alpha $ means more conservative. In other words, the probability of the fire being quicker than $ \zeta^{\alpha} $ is $ \alpha $.} \textcolor{black}{As such, we need to derive an $\alpha$-dependent probability for our upper-bound service time $T_{UB}$, such that it is lower-bounded by the \textit{actual maximum time, $T^*$, a \textcolor{black}{UAV} can go without visiting a $q \in \{Q\}^{N_q}$ and not losing its track quality}. \textcolor{black}{Accordingly,} \textcolor{black}{and} assuming conditional independency between firespots' velocities given latent fire dynamics\footnote{We \textcolor{black}{explicitly estimate the latent fire dynamics} in our AEKF model.}, the probability that our bounds are ``correct'' \textcolor{black}{(meaning that $T_{UB}\geq T^*$)} can be shown as in Equation~\ref{eq:P_Correct}. For our experiments, we set $\alpha=0.05$.} While, specific assumptions made for each Case are discussed in the respective Section, we note that the accuracy of the presented bounds depend on the level of accuracy of the utilized model for fire propagation\footnote{Here, we used the FARSITE model which can be replaced with any other parameterized fire propagation model, such as the correctable fire simulation model introduced in~\cite{beachly2018fire}.}\textcolor{black}{, or generally for applications other than wildfire monitoring, the accuracy of the approximate motion model used for a moving target subject to monitoring}.
\begin{equation}
Pr[{T_{UB} \geq T^*}] \leq 1- (1-\alpha)^{N_q} \label{eq:P_Correct}
\end{equation}

\noindent\textbf{Case 1: Stationary Target Points} -- 	In the first scenario \textcolor{black}{all firespots are} almost stationary \textcolor{black}{(i.e., stationary costumers in the travelling salesman problem). In this case, the search graph, $G_t'$, is a static graph, and therefore, we utilize a minimum spanning tree (MST) to obtain the initial path, $\mathcal{P}_t^{g'}$. The} MST path (i.e., the Hamiltonian cycle computed from the MST) is not the optimal path, but it is fast to generate, \textcolor{black}{and thus, it ensures a quick, sound way to obtain a path that leads to guaranteed service.} Once \textcolor{black}{an initial consensus is reached upon} \textcolor{black}{providing the updated firefront state estimates according to the MST path, the assigned UAVs can then} spend time improving the efficiency of their tours through the $k$-opt \textcolor{black}{algorithm. Accordingly,} we derive the upper-bound time \textcolor{black}{for a \textcolor{black}{UAV} with maximum linear velocity $ v_{max}^d $} to complete a tour \textcolor{black}{on the static search graph in Case 1}, as the Hamiltonian \textcolor{black}{cycle of the generated} MST path, \textcolor{black}{shown by its} \textcolor{black}{spatial} cost \textcolor{black}{$\Delta\mathcal{T}^{\text{MST}}$ as} \textcolor{black}{introduced in Proposition~\ref{prop:case1}}. \textcolor{black}{The $\Delta\mathcal{T}^{\text{MST}}$ is the length of the generated MST path.}

\begin{prop}
\label{prop:case1}
\textcolor{black}{
The upper-bound time for a UAV to complete a monitoring tour, $T_{UB}^{(\text{C1})}$, in Case 1, the stationary target points, can be calculated as shown in Equation~\ref{eq:T_UBcase1}.}
\textcolor{black}{\begin{equation}
\label{eq:T_UBcase1}
T_{UB}^{(\text{C1})} = \frac{2}{v_{max}^d}\Delta\mathcal{T}^{\text{MST}}
\end{equation}}
\end{prop}

Case 1 is presented to develop a tight bound employed under nominal fire conditions and a foundation to derive the traverse temporal upper-bounds for the other cases. Moreover, while there are many approaches for approximating the TSP with guarantees, such as Christofides ($ O(n^3) $), we choose other approaches with lower complexity, such as the double tree ($ O(n^2) $) algorithm. This is because we consider large-scale scenarios with thousands of discrete firespots (discretization of hundreds of square-miles), which signifies the importance of the lower upper-bound on complexity.

\vspace*{0.35cm}
\noindent\textbf{Case 2: Moving Target Points} -- \textcolor{black}{In the second scenario firespots, move but do} not grow (i.e., spawn) considerably. \textcolor{black}{As shown in the middle plot in Figure~\ref{fig:WildfireScenarios}, the upper-bound service time, $T_{UB}$, for a \textcolor{black}{UAV} \textcolor{black}{now} depends not only on the maximum linear velocity of the \textcolor{black}{UAV}, $v_{max}^d$, and the number of nodes $ N_q $ (i.e., length of discretized firespot list $ \{Q\}^{N_q} $), but it also depends on the velocity of the propagating firespots. We derive the} upper-bound on the time \textcolor{black}{it takes a \textcolor{black}{UAV}} to revisit each \textcolor{black}{firespot} location \textcolor{black}{(i.e., each node of the search graph)} \textcolor{black}{for Case 2 as introduced in Proposition~\ref{prop:case2}.} In Equation~\ref{eq:T_UBcase2}, $\left.\reallywidehat{\dot{q}_t^\xi}\right|_{\alpha}$ is the \textcolor{black}{estimated linear velocity of the spot $q_t$} in the $\xi\in\{x,y\}$ direction evaluated at confidence interval defined by $\alpha$.

\begin{prop}
\label{prop:case2}
\textcolor{black}{
The upper-bound time for a UAV to complete a monitoring tour, $T_{UB}^{(\text{C2})}$, in Case 2, the moving target points, can be calculated as shown in Equation~\ref{eq:T_UBcase2}.}
\textcolor{black}{
\begin{align}
T_{UB}^{(\text{C2})} &= \frac{8\zeta^{\alpha}\left(N_q - 1\right)\Delta\mathcal{T}^{\text{MST}}}{v_{max}^d\left(1 - 4\zeta^{\alpha}\left(N_q - 1\right)\right)} \label{eq:T_UBcase2}
\end{align}}
\end{prop}

In Equation~\ref{eq:T_UBcase2}, $\zeta^{\alpha}$ is defined as in Equation~\ref{eq:zetazeta} and we assume a worst-case \textcolor{black}{velocity for all firespots} propagating at the fastest fire's rate of speed in the x- and y-directions, as represented by $\zeta^{\alpha}$. \textcolor{black}{Tuning $\alpha$ enables} control of the degree of confidence in our system at the cost of making the UAV coordination problem more difficult. We emphasis that \textcolor{black}{our} independence and uniformity of fastest and universal \textcolor{black}{velocity assumptions for firespots} pose a worst-case \textcolor{black}{scenario \textcolor{black}{assumption. In other words,} the actual service time, $T^*$, will not become greater than our upper-bound time, $T_{UB}$, if these assumptions are relaxed.}
\begin{equation}
    \zeta^{\alpha} = \argmax_{q,q'}\sqrt{\left(\left.\reallywidehat{\dot{q}_t^x}\right|_{\alpha}\right)^2 + \left(\left.\reallywidehat{\dot{q}_t^y}\right|_{\alpha}\right)^2}
    \label{eq:zetazeta}
\end{equation}

\begin{proof}
To arrive at the bound in Equation~\ref{eq:T_UBcase2}, \textcolor{black}{we start by considering the temporal} cost of traveling to each \textcolor{black}{firespot} under the stationary case, $T_{UB}^{(C1)}$. \textcolor{black}{Moving firespots (even with approximately similar linear velocities), will lead to graph nodes moving possibly in different directions and therefore, the edges of the initial MST graph will shrink or expand. As demonstrated in the middle plot in Figure~\ref{fig:WildfireScenarios}, although the blue arrows, representing velocity and direction of each moving node, have the same lengths (i.e., equal velocities), the next-step tour of the UAV (the dashed black line) is longer than the current tour (solid black line).} In the worst case \textcolor{black}{of firespots moving in opposite directions}, each edge \textcolor{black}{of the search graph} expands according to two times the velocity of the fastest \textcolor{black}{firespot}, $2\zeta^{\alpha}$. Also, if the \textcolor{black}{graph has $N_q$ nodes for the \textcolor{black}{UAV} to consider, this yields $(N_q-1)$ MST edges and $2(N_q-1)$ Hamiltonian paths in-between the nodes.} We note that the total expansion \textcolor{black}{(shrinkage) of the graph} is a function of the time the fire is able to expand \textcolor{black}{(shrink)}, which is a self-referencing \textcolor{black}{relation}, as shown in Equation~\ref{eq:T_UBcase2_start}. \textcolor{black}{Next, by factoring in the universal velocity of the nodes, $\zeta^\alpha$, and replacing $T_{UB}^{(C1)}$ in Equation~\ref{eq:T_UBcase2_start} by its value in Equation~\ref{eq:T_UBcase1} and then solving for $T_{UB}^{(\text{C2})}$, we arrive at Equation~\ref{eq:T_UBcase2}, as demonstrated below.}
\textcolor{black}{
\begin{align}
T_{UB}^{(\text{C2})} &= 4\zeta^{\alpha}\left(N_q - 1\right)\left(T_{UB}^{(\text{C1})} + T_{UB}^{(\text{C2})}\right) \label{eq:T_UBcase2_start} \\
T_{UB}^{(\text{C2})} & \left(1 - 4\zeta^{\alpha}\left(N_q - 1\right)\right) = 4\zeta^{\alpha}\left(N_q - 1\right)T_{UB}^{(\text{C1})} \label{eq:T_UBcase2_start1} \\
T_{UB}^{(\text{C2})} & \left(1 - 4\zeta^{\alpha}\left(N_q - 1\right)\right) = \frac{8\zeta^{\alpha}\left(N_q - 1\right)}{v_{max}^d}\Delta\mathcal{T}^{\text{MST}} \label{eq:T_UBcase2_start2} \\
T_{UB}^{(\text{C2})} &= \frac{8\zeta^{\alpha}\left(N_q - 1\right)\Delta\mathcal{T}^{\text{MST}}}{v_{max}^d\left(1 - 4\zeta^{\alpha}\left(N_q - 1\right)\right)} \label{eq:T_UBcase2_start3}
\end{align}}
\end{proof}

We note that in Case 2, the \textcolor{black}{UAV}'s FOV becomes irrelevant as the tour is based on UAVs' inference of the future distribution of the fire, and the location is not growing so that it would escape the FOV after $ T_{UB} $ steps into the future. Moreover, in the aerial firefront tracking application, when a fire moves to an area that is already burnt or otherwise is observed to dissipate, the fire is pruned from consideration, and the UAV path is updated as there is no fuel for the fire to actually move there.

\vspace*{0.35cm}
\noindent\textbf{Case 3: Moving-Spreading Target Points} -- \textcolor{black}{In the third scenario that we consider, the firespots move and grow quickly, \textcolor{black}{as in a propagating and spreading wildfire}}. \textcolor{black}{In this case,} Single nodes of fire expand over time and escape the \textcolor{black}{UAV}'s FOV. \textcolor{black}{As such, we must consider} the time it takes for a spawning point, \textcolor{black}{$ q_t $, to grow large enough} to escape \textcolor{black}{a \textcolor{black}{UAV}'s FOV}, given \textcolor{black}{the maximum leaner velocity, $v_{max}^d$ and current FOV width, $ w_t $, of the \textcolor{black}{UAV}.} The \textcolor{black}{FOV width} for a \textcolor{black}{UAV} \textcolor{black}{is directly related to the \textcolor{black}{UAV}'s current altitude, $ p_t^z $, and the camera half-angle, $ \phi $, and can be calculated as $w_t = 2p_t^z\tan\phi$.} \textcolor{black}{We derive} the upper bound for \textcolor{black}{\textcolor{black}{UAV}'s} traversal time allowed to maintain the track quality of each \textcolor{black}{firespot} in Case 3 \textcolor{black}{as introduced in Proposition~\ref{prop:case3}.} Equation~\ref{eq:T_UBCase3}.

\begin{prop}
\label{prop:case3}
\textcolor{black}{
The upper-bound time for a UAV to complete a monitoring tour, $T_{UB}^{(\text{C3})}$, in Case 3, the moving-spreading target points, can be calculated as shown in Equation~\ref{eq:T_UBCase3}.}
\textcolor{black}{
\begin{align}
T_{UB}^{(C3)} = \frac{\mathcal{V} + \text{sqrt}\left(\left(1 - \mathcal{V}\right)^2 - \frac{64\mathcal{V}(N_q-1)\Delta\mathcal{T}^{\text{MST}}(\zeta^{\alpha})^2}{v_{max}^{d}w_t\left(1 - 4\zeta^{\alpha}\left(N_q - 1\right)\right)}\right) -1}{4\mathcal{V}\zeta^\alpha (w_t)^{-1}}
\label{eq:T_UBCase3}
\end{align}}
\end{prop}

\textcolor{black}{We refer to the term, $\mathcal{V}$, in Equation~\ref{eq:T_UBCase3} as the \textit{velocity ratio constant}, which captures the ratio between the velocities of the firespots and the \textcolor{black}{UAV} and is computed $\mathcal{V} = \frac{2\zeta^{\alpha}N_q}{v_{max}^d}$.} 

\begin{proof}
To arrive at the bound for Case 3 as in Equation~\ref{eq:T_UBCase3}, \textcolor{black}{we start by quantifying the maximum planar width and length of the enlarging area.} For a specific \textcolor{black}{firespot}, $q_t$, the time-varying \textcolor{black}{planar} width and \textcolor{black}{length} of the enlarging area along \textit{X} and \textit{Y} axes can be \textcolor{black}{computed} as $\mathcal{W}(t) \leq ~2\left.\reallywidehat{\dot{q}_t^x}\right|_{\alpha}T_{UB}^{(C3)}$ and $	\mathcal{L}(t) \leq 2\left.\reallywidehat{\dot{q}_t^y}\right|_{\alpha}T_{UB}^{(C3)}$ at the $\alpha$ confidence level, as \textcolor{black}{firespots} are now allowed to spread for \textcolor{black}{a maximum of} $T_{UB}^{(\text{C3})}$ units of time \textcolor{black}{before the \textcolor{black}{UAV} revisits them}.	Assuming a vertical scanning pattern, we round up the maximum possible $ \mathcal{W}(t) $ and thus, the total number of passes the \textcolor{black}{UAV} would \textcolor{black}{need to} take from left to right \textcolor{black}{can be calculated as} $n(t) \leq \ceil[\Bigg]{ \frac{2\left.\reallywidehat{\dot{q}_t^x}\right|_{\alpha}T_{UB}^{(C3)}}{w_t}}$, and the total distance traversed for each pass is \textcolor{black}{obtained by} $ d(t) = 2\left.\reallywidehat{\dot{q}_t^y}\right|_{\alpha}T_{UB}^{(C3)} $. \textcolor{black}{Accordingly,} the total pass traversed is \textcolor{black}{given by} \textcolor{black}{$ d^{tot}(t) = n(t)d(t) $}. \textcolor{black}{Therefore,} the time \textcolor{black}{it takes} one \textcolor{black}{firespot} to escape \textcolor{black}{the FOV of the \textcolor{black}{UAV}} can be calculated as in Equation~\ref{eq:time_required_final}.
\begin{align}
\tau_q(t) = & \frac{d^{tot}(t)}{v_{max}^d} = \frac{n(t)d(t)}{v_{max}^d} = \frac{\ceil[\Bigg]{ \frac{2\left.\reallywidehat{\dot{q}_t^x}\right|_{\alpha}T_{UB}^{(C3)}}{w_t}}2\left.\reallywidehat{\dot{q}_t^y}\right|_{\alpha}T_{UB}^{(C3)}}{v_{max}^d} \label{eq:time_required_final}
\end{align}\textcolor{black}{\textcolor{black}{In order} to account for all of the firespots, we compute} the summation over all $ \tau_q $ in Equation~\ref{eq:time_required_final}. However, the center of each spreading fire point also moves; \textcolor{black}{thus, $ T_{UB}^{(\text{C2})} $ \textcolor{black}{from} Equation~\ref{eq:T_UBcase2} must also} be added to this summation, \textcolor{black}{as shown in Equation~\ref{eq:T_UBCase3_start_appx}.}
\begin{align}
T_{UB}^{(C3)} &=  T_{UB}^{(\text{C2})}+  \sum_{q\in\{Q\}} \frac{2}{v_{max}^d}\left.\reallywidehat{\dot{q}_t^y}\right|_{\alpha}T_{UB} \ceil[\Bigg]{ \frac{2\left.\reallywidehat{\dot{q}_t^x}\right|_{\alpha}T_{UB}^{(C3)}}{w_t}} \label{eq:T_UBCase3_start_appx}
\end{align}To solve Equation~\ref{eq:T_UBCase3_start_appx} for $ T_{UB}^{(C3)} $ and find the final upper-bound, we \textcolor{black}{make two} simplifying assumptions. First, we remove the ceiling operator and add one to the term inside the operator to achieve continuity, as shown in Equation~\ref{eq:ceil}. 
\begin{align}
\ceil[\Bigg]{ \frac{2\left.\reallywidehat{\dot{q}_t^x}\right|_{\alpha}T_{UB}^{(C3)}}{w_t}}\leq { \frac{2\left.\reallywidehat{\dot{q}_t^x}\right|_{\alpha}T_{UB}^{(C3)}}{w_t}} + 1 \label{eq:ceil}
\end{align}Second, we adopt a similar approach to Case 2 by assuming the area required to cover to account for the growth of each fire location is upper-bounded by $N_q$ times the area of growth for a hypothetical fire growing quickest (Equation~\ref{eq:fastest}).
\begin{align}
&\sum_q \left.\reallywidehat{\dot{q}_t^y}\right|_{\alpha}T_{UB} \ceil[\Bigg]{ \frac{2\left.\reallywidehat{\dot{q}_t^x}\right|_{\alpha}T_{UB}^{(C3)}}{w_t}} \leq  N_q\zeta^{\alpha} \ceil[\Bigg]{ \frac{2\zeta^{\alpha}T_{UB}^{(C3)}}{w_t}} \label{eq:fastest}
\end{align}With these two conservative assumptions and replacing $ T_{UB}^{(\text{C2})} $ from Equation~\ref{eq:T_UBcase2}, \textcolor{black}{the upper-bound service time} in Equation~\ref{eq:T_UBCase3_start_appx} can be revised as \textcolor{black}{shown below, in Equation~\ref{eq:T_UBCase3_start_appx_revised}-\ref{eq:T_UBCase3_start_appx_revised1}.}
\begin{align}
T_{UB}^{(C3)} &=  T_{UB}^{(\text{C2})} +  \frac{2N_q\zeta^{\alpha}T_{UB}^{(C3)}}{v_{max}^d}\left[ \frac{2\zeta^{\alpha}T_{UB}^{(C3)}}{w_t} +1 \right]  \label{eq:T_UBCase3_start_appx_revised} \\ 
T_{UB}^{(C3)} &- \frac{2N_q\zeta^{\alpha}T_{UB}^{(C3)}}{v_{max}^d}\left[ \frac{2\zeta^{\alpha}T_{UB}^{(C3)}}{w_t} +1 \right] = \frac{8\zeta^{\alpha}\left(N_q - 1\right)\Delta\mathcal{T}^{\text{MST}}}{v_{max}^d\left(1 - 4\zeta^{\alpha}\left(N_q - 1\right)\right)} \label{eq:T_UBCase3_start_appx_revised1}
\end{align}We note that Equation~\ref{eq:T_UBCase3_start_appx_revised1} is in the form of a general quadratic equations \textcolor{black}{(i.e., $ z-az\left(bz+1\right) =\delta $)} \textcolor{black}{in which} $ z = T_{UB}^{(C3)} $. \textcolor{black}{This form can be reorganized into $ \gamma z^2 - \beta z + \delta = 0 $, where $ \gamma = ab = \frac{4N_q\left(\zeta^{\alpha}\right)^2}{w_tv_{max}^d} $ and, $ \beta = 1-a = 1 - \frac{2N_q\zeta^{\alpha}}{v_{max}^d} $ and,} $\delta = \frac{8\zeta^{\alpha}\left(N_q - 1\right)\Delta\mathcal{T}^{\text{MST}}}{v_{max}^d\left(1 - 4\zeta^{\alpha}\left(N_q - 1\right)\right)}$. \textcolor{black}{The upper-bound traversal time, $ T_{UB}^{(C3)} $, for Case 3 can be obtained from the general form of solutions to quadratic equations, $T_{UB}^{(C3)} = \frac{-\beta + \sqrt{\beta^2 - 4\gamma\delta}}{2\gamma}$,} in which replacing $\gamma$, $\beta$ and $\delta$ results in Equation~\ref{eq:T_UBCase3_app}. \textcolor{black}{Finally, plugging in the velocity ratio constant, $\mathcal{V}$, into the Equation~\ref{eq:T_UBCase3_app} and rearranging the terms obtains $T_{UB}^{(C3)}$ as presented in Equation~\ref{eq:T_UBCase3}.}
\textcolor{black}{
\begin{align}
T_{UB}^{(C3)} = \frac{\left(\frac{2N_q\zeta^{\alpha}}{v_{max}^d}-1 + \sqrt{\left(1 - \frac{2N_q\zeta^{\alpha}}{v_{max}^d}\right)^2 - \frac{128N_q(N_q-1)\Delta\mathcal{T}^{\text{MST}}\left(\zeta^{\alpha}\right)^3}{w_t(v_{max}^d)^2\left(1 - 4\zeta^{\alpha}\left(N_q - 1\right)\right)}}\right)}{8(w_tv_{max}^d)^{-1}N_q\left(\zeta^{\alpha}\right)^2}
	\label{eq:T_UBCase3_app}
\end{align}}
\end{proof}

\section{Coordinated Distributed Field Coverage Module (No Specified Areas)}
\label{subsec:distributedcoverage}
\noindent UAV agents in the team which are not assigned to monitor the firefronts within \textcolor{black}{specified (human-defined) areas of priority} directly \textcolor{black}{(i.e., unallocated UAVs)}, can be used to explore the rest of the wildfire. See Figure~\ref{fig:SafeHRI} as an example illustration. \textcolor{black}{When we have access to additional UAV resources that may not be needed for the performance-guaranteed coverage of the prioritized areas, we can deploy such UAVs (i.e., unallocated UAVs) to monitor the rest of the wildfire areas.} \textcolor{black}{To this end, \textcolor{black}{in this section} we propose a coordinated, distributed field coverage framework for multiple UAVs to collectively cover and monitor the rest of wildfire areas, other than areas specified by human-teams}. \textcolor{black}{We show that how as a corollary of deriving our analytical upper-bound service times, $T_{UB}$, these temporal values can be leveraged to design a simple yet effective coverage strategy.} 

To design our proposed coordinated, collaborative field coverage algorithm we focus on settings with \textit{Centralized Planning and Distributed Execution} (CPDE) \textcolor{black}{paradigm}. In other words, communication between UAV agents is not restricted during the planning phase which is done by a centralized computer; however, during execution of the assigned plans, each UAV only performs locally and can communicate with its neighboring UAVs. \textcolor{black}{We note that the CPDE paradigm} is a standard and widely-used setting for multi-agent planning~\cite{kraemer2016multi,foerster2016learning,seraj2020adaptive}. The steps to our \textcolor{black}{distributed} coverage module \textcolor{black}{for non-prioritized areas} are summarized in \textcolor{black}{Algorithm~\ref{alg:method1}} and are \textcolor{black}{elaborated} here.

\begin{algorithm}
	\footnotesize
	\SetKwData{NumUAV}{NumUAV}\SetKwData{TimeAvailable}{TimeAvailable}\SetKwData{MissionDuration}{MissionDuration}\SetKwData{True}{True}\SetKwData{False}{False}
	\SetKwFunction{CETSP}{CETSP}\SetKwFunction{UpdateDroneStats}{UpdateDroneStats}\SetKwFunction{UpdateMap}{UpdateMap}\SetKwFunction{CSP}{CSP}\SetKwFunction{Sense}{Sense}\SetKwFunction{ComputePath}{ComputePath}\SetKwFunction{FeasibilityTest}{FeasibilityTest}\SetKwFunction{Cluster}{Cluster}\SetKwFunction{TUB}{TUB}\SetKwFunction{CoordinatedCoverage}{CoordinatedCoverage}\SetKwFunction{len}{len}\SetKwFunction{MST}{MST}\SetKwFunction{kopt}{k-opt}
	\SetKwInOut{Input}{input}\SetKwInOut{Output}{output}
	\Input{Obtain the fire-map $ \{Q_t\}^{N_q} $, list of all UAVs $ \{\text{UAV}\}^{N_d} $, set of UAV poses, velocities, etc.}
	\Output{Assigned \textcolor{black}{UAV}-path pairs $ \{\mathcal{A}_t^d\} $}
	
	\textbf{objective:} Unallocated UAVs, $\{\text{UAV}\}^{N_d^0}$, to cover unspecified areas, $\{Q_t\}^{N_q^0}$
	\BlankLine
	
	// main loop //
	
	Initialize $t\leftarrow 0$
	
	\While{MissionDuration}{
		
		Update fire-map and priority areas: $ \langle\{Q_t\}^{N_q^1}, \{Q_t\}^{N_q^0}\rangle \leftarrow $ \UpdateMap{$\{Q_t\}^{N_q}$}
		
		Update UAVs status: $ \langle\{\text{UAV}\}^{N_d^1}, \{\text{UAV}\}^{N_d^0}\rangle \leftarrow $ \UpdateDroneStats{$\{\text{UAV}\}^{N_d}$}
		
		\CoordinatedCoverage($\{Q_t\}^{N_q^0}, \{\text{UAV}\}^{N_d^0}, t$) // pass in the time $t$
		
		Update time: $t\leftarrow t+1$
		
			
			
	}
	// inner functions //
	
	\textbf{def} ~\CoordinatedCoverage($\{Q_t\}^{N_q^0}, \{\text{UAV}\}^{N_d^0}, t$)\textbf{:}
	
	\hspace{0.5cm}$ \langle\{T_{UB_{i1}}^{g'},..,T_{UB_{i}}^{g'}\}, \{\mathcal{P}'_{i1},..,\mathcal{P}'_{i}\}\rangle \leftarrow $ \CSP{\TUB{\Cluster{$\{Q_t\}, \len(\{\text{UAV}\}^{N_d^0})$}}}
	
	\hspace{0.5cm}\textbf{if} $\left(t\geq \min(T_{UB}^i)\right)$ || $\left(i\in\{\text{UAV}\}^{N_d^0} ~\text{is deployed for human support}\right)$
	
	\hspace{1cm}$ \langle\{\mathcal{P}'_{i1},\cdots,\mathcal{P}'_{i_{new}}\}\rangle \leftarrow $ \Cluster{$\{Q_t\}, \len(\{\text{UAV}\}^{N_d^0}_{new})$} // re-clustering
	
	\hspace{1cm} Repeat line 12 for the new assignments
	
	\caption{\textcolor{black}{Stages of the proposed distributed field coverage module when no area is prioritized for surveillance.}}
	
	\label{alg:method1}
	
\end{algorithm}

\textcolor{black}{After detecting the fire map and hotspots, a} set of firespots based on the aforementioned Steiner zone variable neighborhood search method are generated (see Section~\ref{subsubsec:CETSP}). The set of nodes are partitioned (i.e., K-means clustering) according to the number of available (\textcolor{black}{i.e.}, unallocated) UAVs. Each \textcolor{black}{UAV} is assigned to one partition by solving a simple constraint satisfaction problem (CSP) with \textcolor{black}{UAV}s as variables, partitions as domains, and distance to the partition centroid as constraints (Line 12 in Algorithm~\ref{alg:method1}). Note that the partitioning step and solving the CSP problem is done centrally, during the centralized planning phase. 

After assigning UAVs to a partition, each UAV agent starts to execute the following tasks distributedly \textcolor{black}{(Line 12 in Algorithm~\ref{alg:method1})}. First, an optimal path for coverage and tracking is found by applying the k-opt algorithm. The upper-bound time, $ T_{UB} $, is then calculated for a firespot in the center of \textcolor{black}{a} \textcolor{black}{UAV}'s FOV. $T_{UB}$ provides the \textcolor{black}{UAV} with an estimate of how long it will take a fire to escape its FOV as determined by fire propagation velocity. When a route is identified in this way, UAVs can apply this reasoning for the next $ T_{UB} $ time-steps before recalculating a new path \textcolor{black}{(Line 13 in Algorithm~\ref{alg:method1})}. After $T_{UB}$ has passed, the partitions are revised, and an optimal path is recalculated, since fire locations have likely changed significantly during this time \textcolor{black}{(Lines 14-15 in Algorithm~\ref{alg:method1})}. Once a \textcolor{black}{UAV} is \textcolor{black}{recruited for help to monitor firefronts within human-specified areas, one of the unallocated \textcolor{black}{UAV}s} is dispatched and the process is repeated from the central planning phase. \textcolor{black}{In scenarios where the centralization of the planning step does not impose an issue, leveraging this method provides a simple, low-complexity approach that plans for multiple UAVs at the high-level (without considering the low-level control inputs) to efficiently cover a dynamic field.}

We emphasize that the relation between our coordinated, distributed field coverage algorithm described here and the coordinated planning framework (described in Section~\ref{sec:method}) is that the latter coordinates UAV agents to efficiently track the firefronts within \textcolor{black}{specified} disjoint areas of fire (\textcolor{black}{i.e.}, areas prioritized by human firefighters) by using as few UAVs as possible, while guaranteeing the performance. \textcolor{black}{The algorithm described in this section utilizes the remaining UAVs in the team (if any) to collectively surveil the rest of the wildfire area, if needed.} In general, this module may not be needed, or it can be replaced with any other surveillance methods depending on the problem setting (\textcolor{black}{e.g.,} if centralized planning is possible or if it needs to be fully distributed). For instance, if the centralized planning is not feasible in an application, the coverage method in this section can be replaced with the fully decentralized controller in \cite{seraj2020coordinated}, or any other similar methods from the literature.

\section{Empirical Evaluation}
\label{sec:results}
\noindent In this section, we empirically evaluate both our \textcolor{black}{multi-UAV coordination and planning framework (Section~\ref{sec:method})} as well as the coordinated, distributed field coverage module introduced in Section~\ref{subsec:distributedcoverage} in an aerial firefront tracking and wildfire area monitoring case-study. In our experiments, robots in a team of homogeneous, autonomous UAVs (e.g., omni-directional multi-rotor aircrafts such as quadcopters) are tasked to coordinate together to: (1) track and monitor the firefront within \textcolor{black}{multiple specified human-defined vicinities of priority} and provide \textcolor{black}{real-time fire states and tracking information} and (2) cooperatively cover and surveil the entire wildfire area. For the field coverage we test the performance of our approach in comparison with two state-of-the-art model-based and reinforcement learning benchmarks \textcolor{black}{(see Section~\ref{subsec:benchmarks})}. Moreover, we demonstrate the feasibility of our framework through implementation on physical robots in a multi-robot testbed (Section~\ref{sec:experimentalresults}).

\subsection{\textcolor{black}{Baselines}}
\label{subsec:benchmarks}
\noindent The first benchmark to which we compare is \textcolor{black}{a recent} distributed control framework for dynamic wildfire tracking as proposed by \cite{pham2017distributed}. The distributed control framework includes two controller modules in which one is responsible for field coverage and the other for path planning (i.e., in-flight collision avoidance, maintaining a safe altitude, and moving towards new desired poses). The two controllers are defined as the negative gradients (gradient descent) of objective functions to maximize the area-pixel density of the UAV's fire observations and to maintain safe flight parameters with potential field-based criteria, respectively.

Second, we compare our approach to a reinforcement learning (RL) benchmark method proposed in~\cite{haksar2018distributed}. Each UAV is controlled by an independent agent, and all agent policies are identical (scalable RL algorithm). The policy network architecture consists of three fully connected layers with ReLU activations, following prior work~\cite{haksar2018distributed}. The agent receives as input an uncertainty map over the wildfire, as described in Section \ref{subsubsec:URR}, and outputs a direction to move for the next time-step. The reward function is adapted from prior work, though elements that involve collision are removed, as we assume \textcolor{black}{UAV}s can occupy the same space. We follow hyperparameter settings given in prior work, and perform a sweep over hidden-layer dimensions, as they are not explicitly defined in prior work.

\subsection{Simulation Environment and Results}
\label{subsec:results}
\subsubsection{\textcolor{black}{Evaluating the Algorithms}}
\label{subsubsec:EvaluatingtheAlgorithms}
\begin{figure}[t!]
	\centering
	\includegraphics[width=\columnwidth]{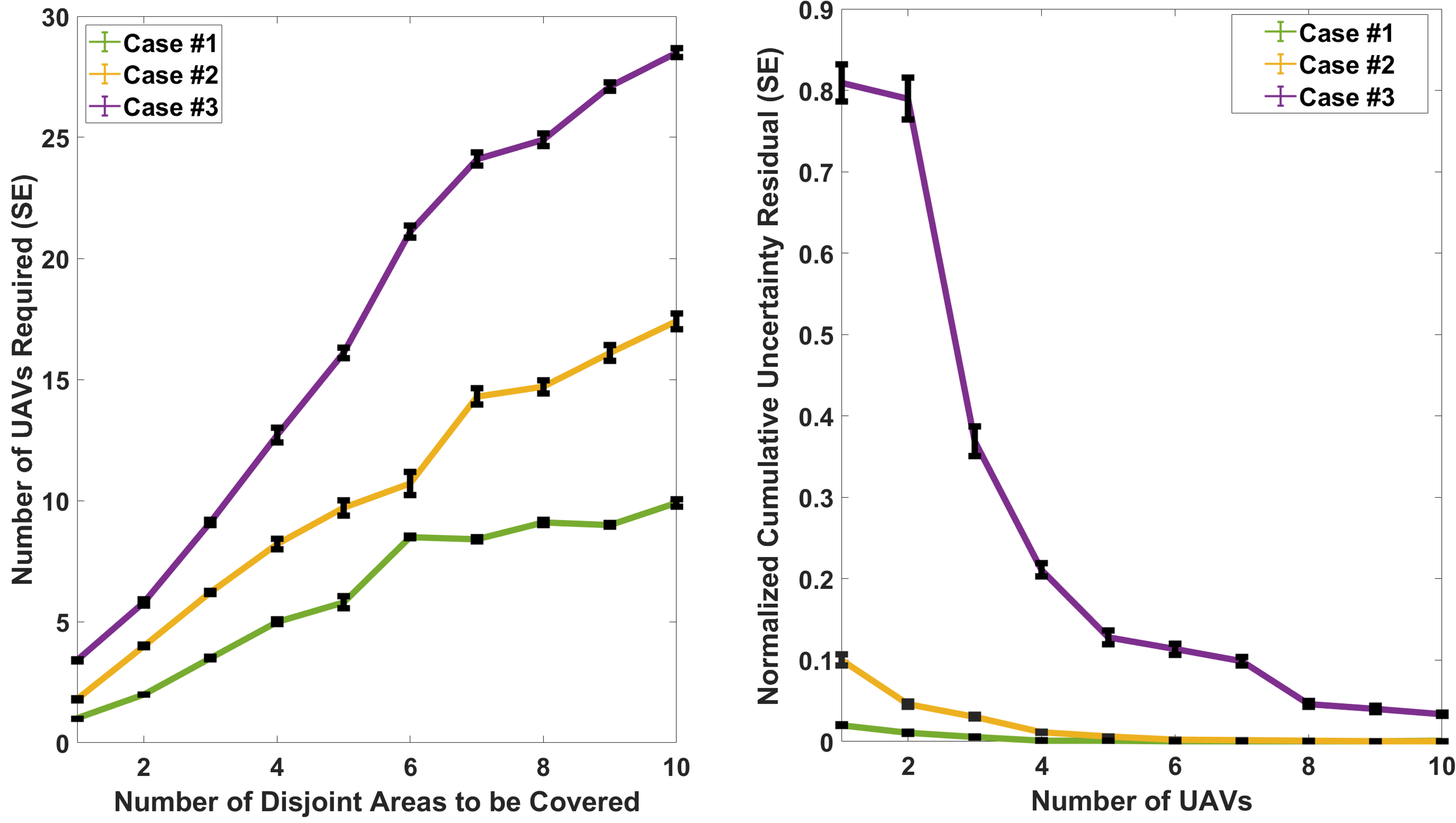}
	\caption{\textcolor{black}{The left-side figure depicts a quantitative evaluation of our analytical URR bound and shows that increasing the number of disjoint areas to be monitored resulted in a rise in the number of \textcolor{black}{UAV}s required to guarantee the performance. This also held as the wildfire propagation scenario changed and a fire propagated more aggressively. The right-side figure demonstrates the efficacy of the coordinated, distributed coverage algorithm, in which less cumulative uncertainty residuals (Y-axis) is better. This figure shows that \textcolor{black}{UAV}s easily covered the fire map in the case of stationary wildfire, while in the case of a fast-growing fire, more \textcolor{black}{UAV}s were required to cover the fire map efficiently.}}
	\label{fig:humanSafety_module_results_all}
	\vspace*{-0.5cm}
\end{figure}

\noindent To evaluate the efficacy of our proposed \textcolor{black}{multi-UAV planning framework for firefront tracking}, we performed a comprehensive simulation to determine the number of \textcolor{black}{UAV}s needed to satisfy the uncertainty residual ratio (given in Equation~\ref{eq:URR1}) for a range of \textcolor{black}{numbers of disjoint areas specified for firefront tracking}. We performed the evaluation for all three mentioned wildfire scenarios for ten trials and calculated the mean and standard error (SE) for each. The objective was to determine the number of required \textcolor{black}{UAV}s to \textcolor{black}{guarantee the performance} in each case. The results for this simulations are presented in Figure~\ref{fig:humanSafety_module_results_all} (left-side figure). 

For the aforementioned evaluations, we simulated one to \textcolor{black}{ten (\textcolor{black}{i.e.}, $N_h\in\{1, 2, .., 10\}$)} distant areas of fire, assuming each corresponds to one prioritized area. Each area included 20-30 (\textcolor{black}{i.e.}, random integer in the range $\left[20, 30\right]$) randomly placed fire-spots. \textcolor{black}{Areas were randomly initialized} within a 500-by-500 terrain. A UAV team was randomly positioned within a distant \textcolor{black}{location at a corner of the terrain}. A total number of 30 \textcolor{black}{UAV}s were assumed as the goal was to determine the number of required \textcolor{black}{UAV}s to \textcolor{black}{guarantee the quality of service} in each case. We chose the fire propagation velocity to be 0, 0.5, and 1 for wildfire Cases 1, 2, and 3, respectively. Moreover, we chose the spawning rate of the fire for Case 3 to be at most three (i.e., each fire can produce up to three more fires). Additionally, the team of UAVs were homogeneous in their dynamics and motion characteristics where the maximum linear velocity, $ v_{max}^d $, of the \textcolor{black}{UAV}s was set to 500 for all \textcolor{black}{UAV}s.

\begin{figure*}[t!]
	\centering
	\includegraphics[width=1.0\columnwidth]{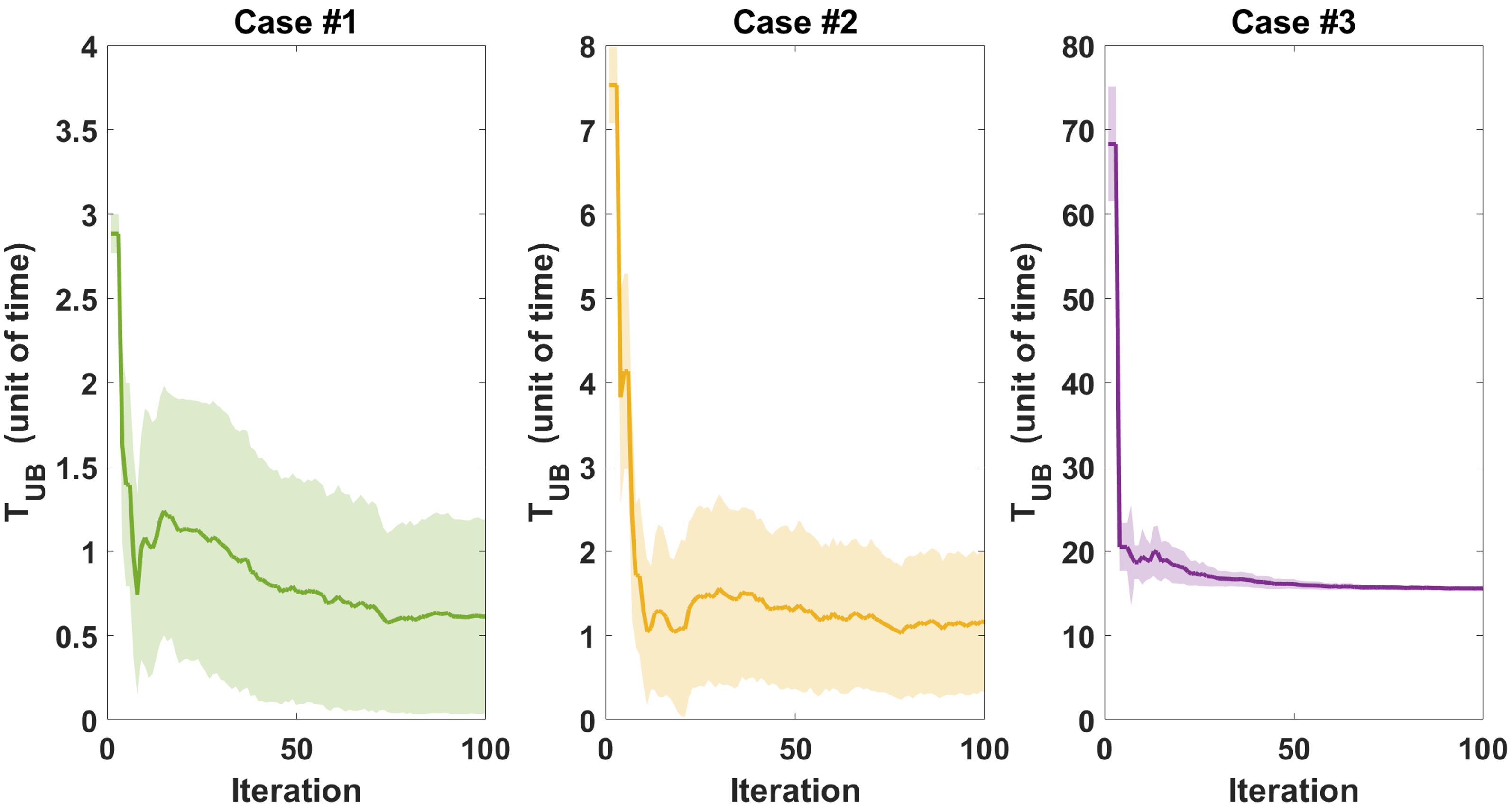}
	\caption{\textcolor{black}{The computed upper-bound times, $T_{UB}$, it takes a UAV to complete a determined tour over the course of a simulation with respect to the three wildfire scenarios. As shown, the computed upper-bound service times initially start with a high value, relative to the wildfire scenario, and gradually decrease until converging to a reasonable value, as the algorithm determines that more UAVs are required for a guaranteed tracking of the firespots. Plots are averaged across UAVs and across ten separate experiment trials.}}
	\label{fig:TUB_All}
	\vspace*{-0.5cm}
\end{figure*}

\textcolor{black}{To further investigate the UAV-team behavior during our multi-UAV planning framework in the above experiment, we demonstrated the computed $T_{UB}$ times by the UAVs in Figure~\ref{fig:TUB_All}. Figure~\ref{fig:TUB_All} shows the computed upper-bound times, $T_{UB}$, that it takes UAVs to complete a determined tour with respect to the three wildfire scenarios. The plots in Figure~\ref{fig:TUB_All} are averaged across UAVs and across ten separate experiment trials. As shown, the computed upper-bound service times initially start with a high value, relative to the wildfire scenario since, at first, only one UAV is assigned to monitor and track the entire firefront points. As the time proceeds and the algorithm determines that more UAVs are required for a guaranteed tracking of the firespots, the computed $T_{UB}$ times become smaller until the values converge to a reasonable time respective to the fire scenario.}

\begin{figure}
    \centering
    \begin{subfigure}[t]{0.32\columnwidth}
        \centering
        \includegraphics[width = \linewidth]{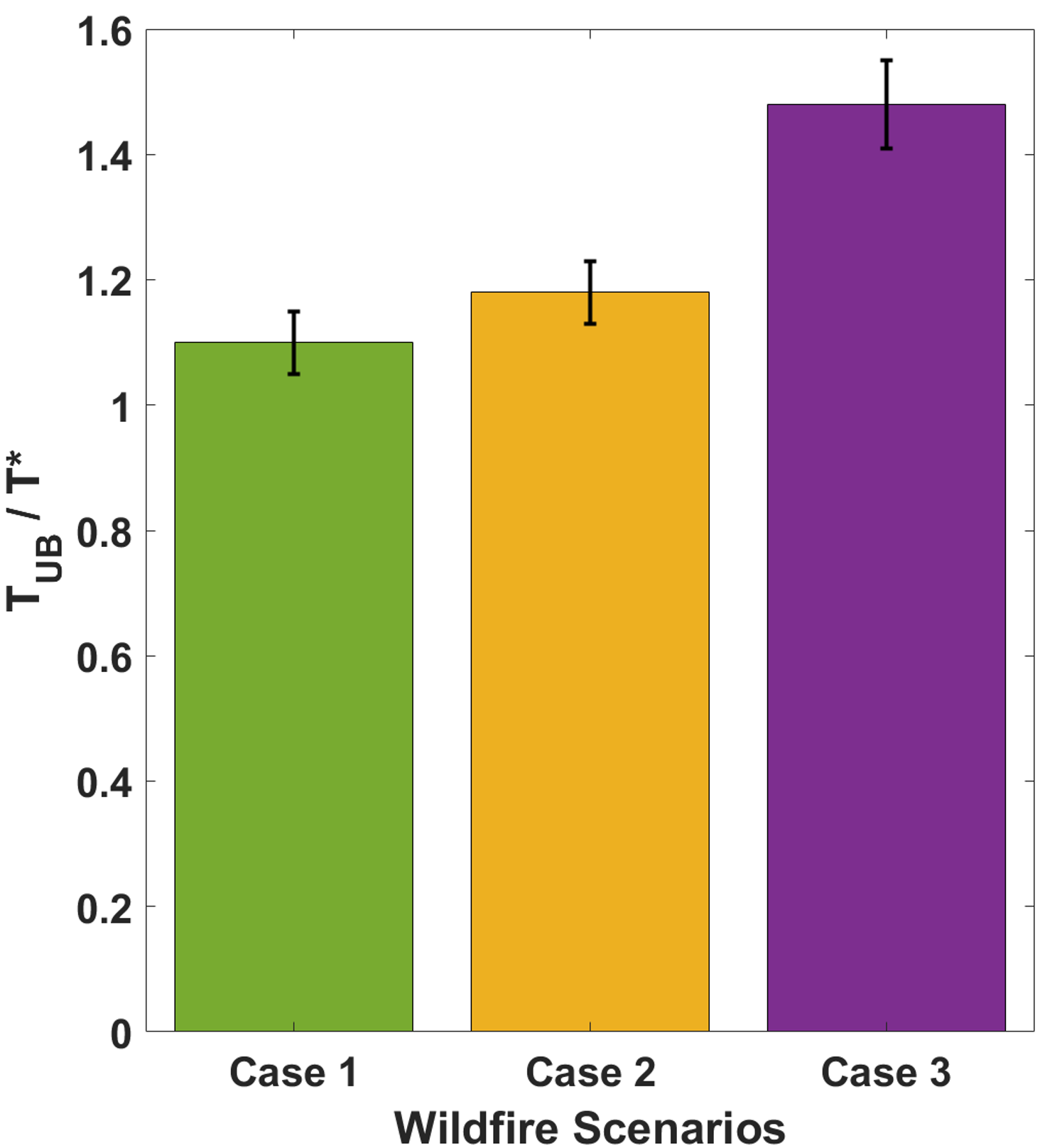}
        \caption{\textcolor{black}{$\frac{T_{UB}}{T^*}$ ratio for all fire scenarios.}}\label{fig:tub_tstar_ratio}
    \end{subfigure}
    \begin{subfigure}[t]{0.32\columnwidth}
        \centering
        \includegraphics[width = \linewidth]{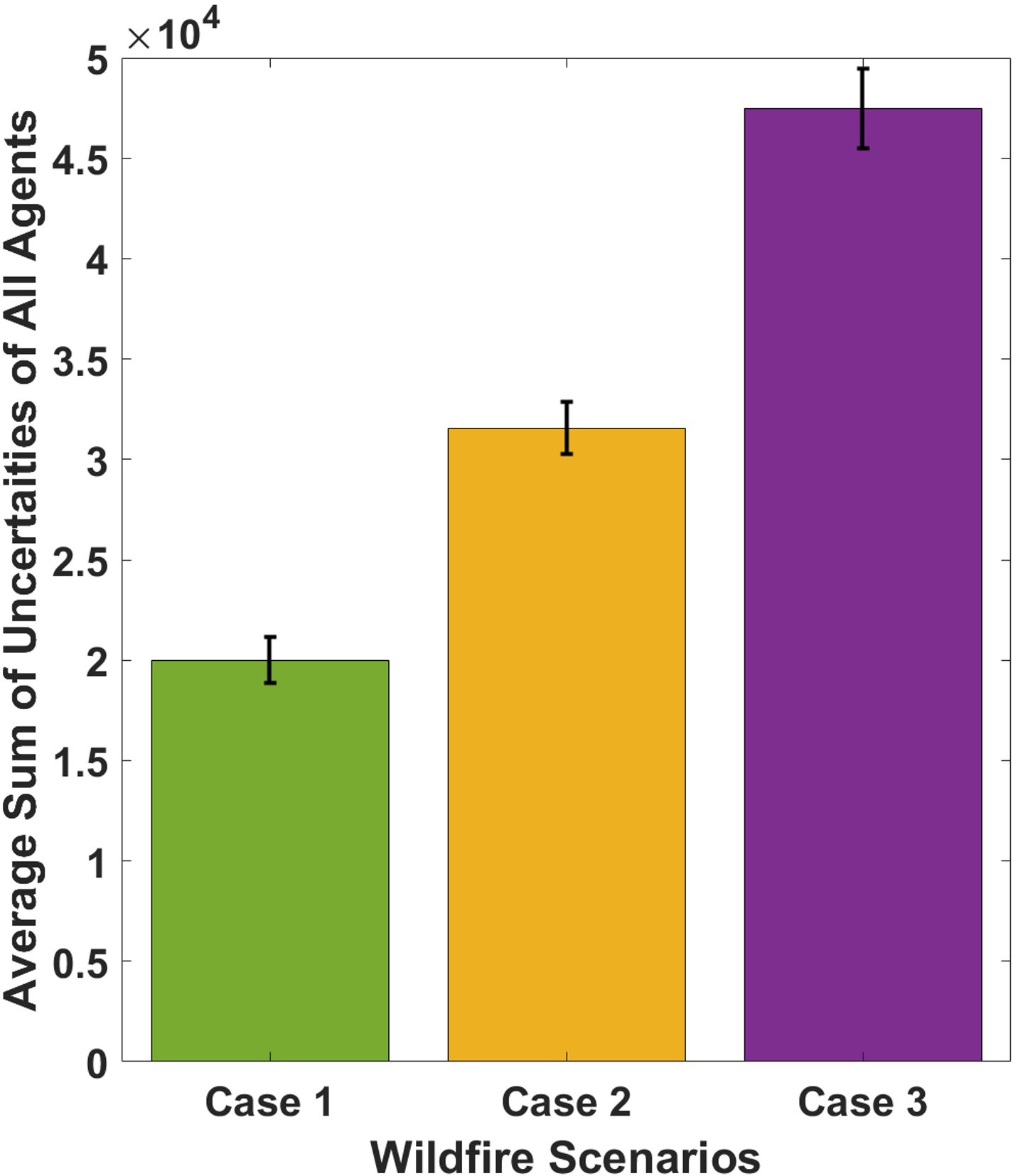}
        \caption{\textcolor{black}{Average sum of all uncertainties over time.}}\label{fig:unc_bargraphs}
    \end{subfigure}
    \begin{subfigure}[t]{0.32\columnwidth}
    \centering
        \includegraphics[width = \linewidth]{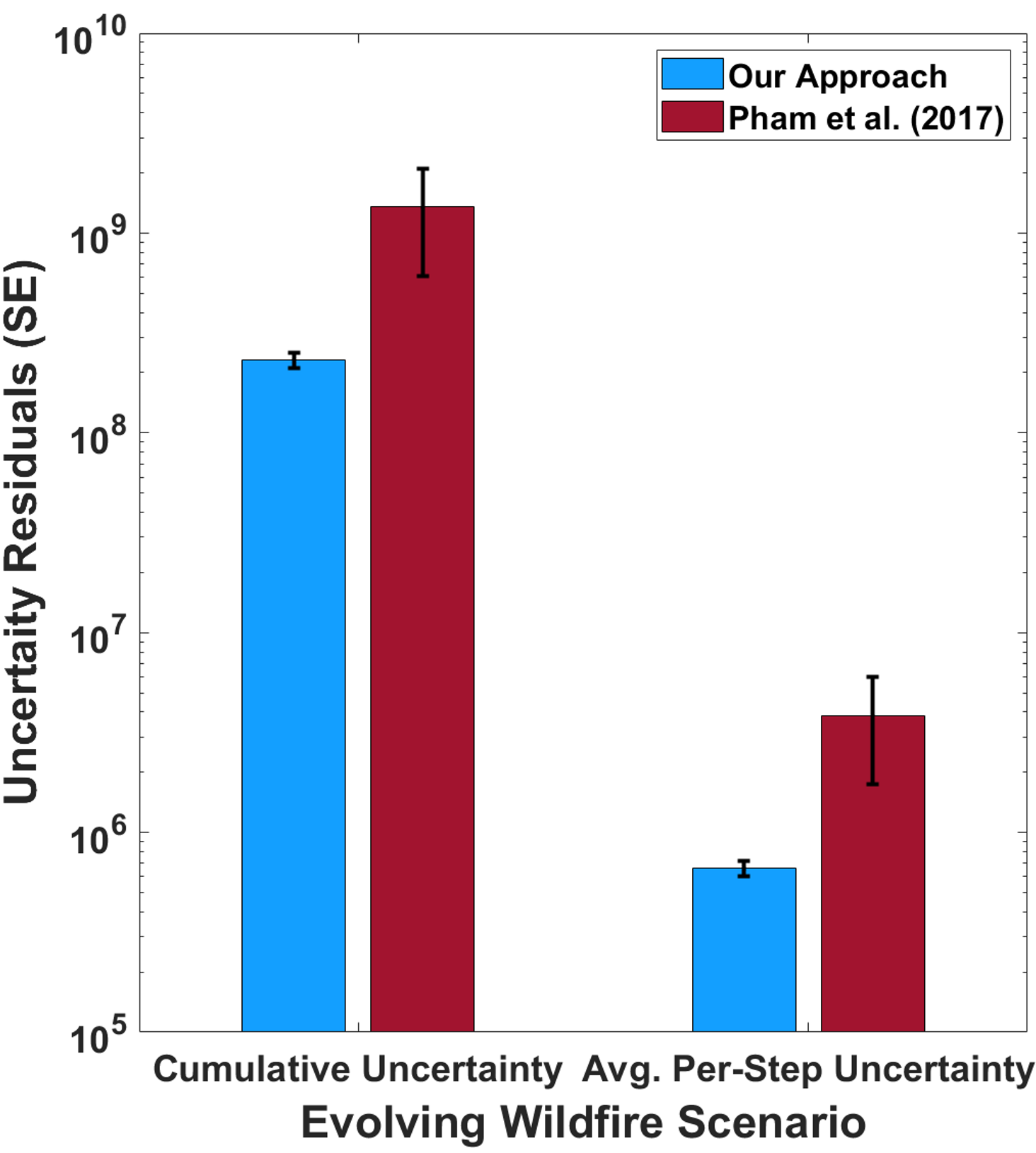}
        \caption{\textcolor{black}{The uncertainty residuals in evolving fire scenario.}}\label{fig:evolving_fire_result}
    \end{subfigure}
    \begin{subfigure}[t]{1.0\columnwidth}
    \centering
        \includegraphics[width = \linewidth]{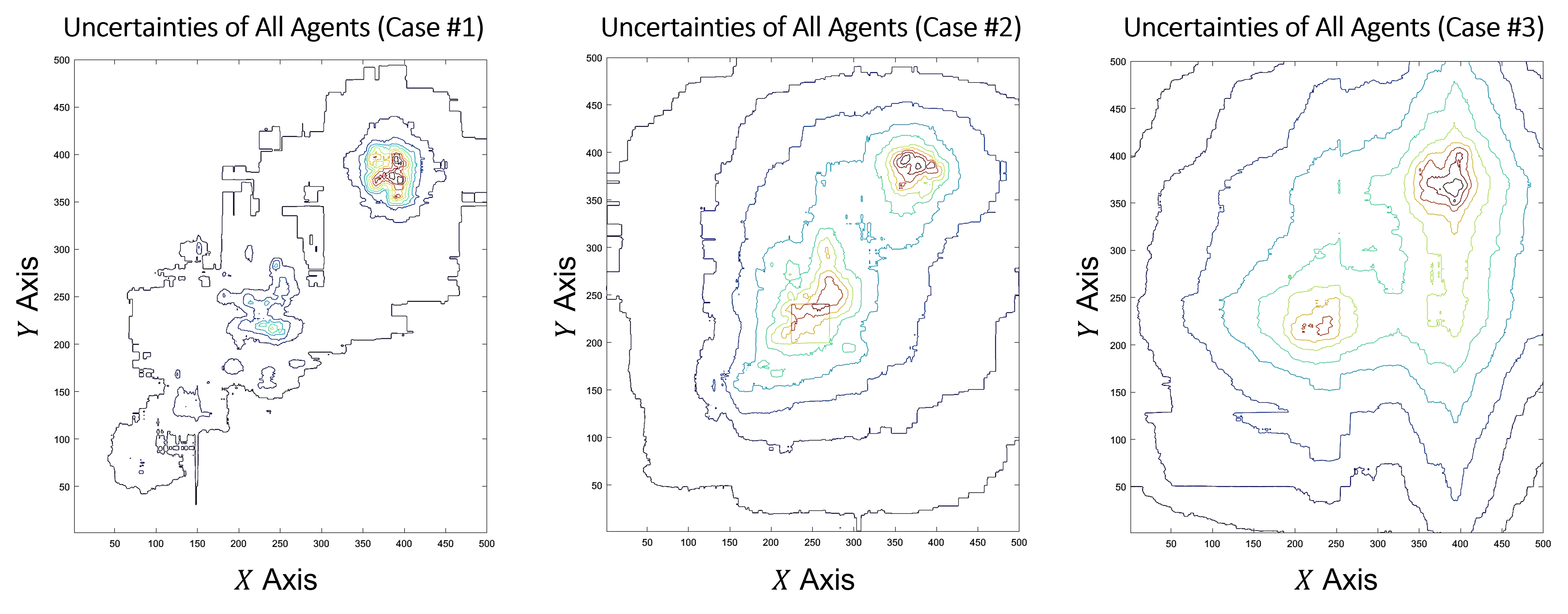}
        \caption{\textcolor{black}{The combined uncertainty maps generated by all UAV agents for all firespots, averaged over time.}}\label{fig:All_Unc}
    \end{subfigure}
    \caption{\textcolor{black}{Figure~\ref{fig:tub_tstar_ratio} shows the $\frac{T_{UB}}{T^*}$ ratio for all fire scenarios computed over 5000 trials for each case. This ratio evaluates the sanity of the derived upper-bound times, $T_{UB}$, and the actual maximum time it takes UAVs to complete a tour, $T^*$. As shown, for a tight upper-bound, the value of this ratio must remain close to one. Figure~\ref{fig:unc_bargraphs} shows the sum of all UAVs' uncertainties for all of the measurements, averaged over time. As shown, the measurement uncertain residuals are bounded. Figure~\ref{fig:evolving_fire_result} compares the measurement uncertainty residuals for our approach and~\cite{pham2017distributed} in the evolving fire scenario (i.e., fire starts in Case 1 and evolves into Case 2 and 3 at certain time-steps). As shown, our method outperforms the baseline in this more challenging scenario. Figure~\ref{fig:All_Unc} shows the combined uncertainty maps generated by summing all measurement residuals of all UAVs for all firespots and averaging over time.}}
    \label{fig:BaselineComp}
    \vspace*{-0.25cm}
\end{figure}

\textcolor{black}{Additionally, to demonstrate that our derived upper-bound times, $T_{UB}$, are tight with respect to the actual maximum time it takes UAVs to complete a tour, $T^*$, we computed the $\frac{T_{UB}}{T^*}$ ratio for all fire scenarios in an experiment. This \textcolor{black}{empirically-evaluated} ratio evaluates the tightness of the derived upper-bound times where for a tight bound, the value of this ratio must remain close to one. Figure~\ref{fig:tub_tstar_ratio} shows the mean value of this ratio ($\pm$ standard error), computed over 5000 trials for each fire scenario. As shown, the value of the ratio slightly increases as the fire scenario gets more intense. The upper-bounds for Case 1 (i.e., Equation~\ref{eq:T_UBcase1}), Case 2 (i.e., Equation~\ref{eq:T_UBcase2}), and Case 3, (i.e., Equation~\ref{eq:T_UBCase3}) are 1.1$\times$, 1.18$\times$, and 1.48$\times$ greater than the actual maximum times, $T^*$, for the respective cases. \textcolor{black}{Results show that our} upper-bounds become more conservative for more aggressive fire scenarios. We note that for this experiment, the environment parameters are as described above, in Section~\ref{subsubsec:EvaluatingtheAlgorithms}, and the actual maximum times, $T^*$, were computed directly in simulation and without considering the MST paths designed in our algorithm.}

We also evaluated our \textcolor{black}{coordinated, distributed field coverage} algorithm within a similar framework by calculating the cumulative uncertainty residual while tasking an increasing number of UAVs to surveil a large propagating wildfire map \textcolor{black}{(\textcolor{black}{i.e.}, no specified areas)}. The uncertainty residual at each \textcolor{black}{timestep} was calculated by inspecting the ground-truth fire map for firespots that were not covered by any \textcolor{black}{UAV}s, and the respective cumulative error residual obtained from AEKF was calculated. The uncertainty residual measure was an indicator of how successful the team of UAVs were in cooperatively covering all firespots and increased by the number of nodes not covered by any \textcolor{black}{UAV}. Figure~\ref{fig:humanSafety_module_results_all} (right-side) shows the results for the evaluation of our \textcolor{black}{coordinated, distributed coverage algorithm.} All wildfire environment and UAV characteristics and parameters for experiments in Figure~\ref{fig:humanSafety_module_results_all} (right-side) were similar to the description presented above \textcolor{black}{and once again, the experiments were conducted for all three aforementioned wildfire scenarios.}

\textcolor{black}{We further investigated the efficacy of our algorithm by: (1) assessing the boundedness of the UAVs' measurement residuals through cumulative uncertainties and (2) evaluating the performance in an evolving wildfire scenario. For these experiments, we initialized two distinct fire areas in a large 500$\times$500 terrain, each with around 30 initial firespots. Figure~\ref{fig:unc_bargraphs} shows the sum of all agents' uncertainties for all measurements, averaged over time and across ten trials. As shown, the measurement uncertain residuals are bounded while the values increase as the wildfire scenario becomes more aggressive. Figure~\ref{fig:All_Unc} illustrates the combined uncertainty maps in the 500$\times$500 terrain for each of the three wildfire scenarios by summing all the measurement residuals of all agents for all firespots and averaging over time. Figure~\ref{fig:evolving_fire_result} compares the measurement uncertainty residuals for our coordinated field coverage approach and~\cite{pham2017distributed} in an evolving fire scenario. For this experiment, firespots start in Case 1 (i.e., stationary) and evolve into Case 2 and 3 at $t=50$ and $t=150$, respectively for a total of $t=350$ time-steps. A total of five UAVs initiated at a distant location are tasked to cover and track the wildfire area at all times. As shown \textcolor{black}{in Figure~\ref{fig:evolving_fire_result}}, our method outperforms the baseline in this challenging scenario by accumulating $5.8\times$ less uncertainty residual than the baseline over the simulation time and on average per time-step.}

\subsubsection{\textcolor{black}{Baseline Comparison}}
\label{subsubsec:Baseline Comparison}
\begin{figure*}[t!]
	\centering
	\includegraphics[width=1.0\columnwidth]{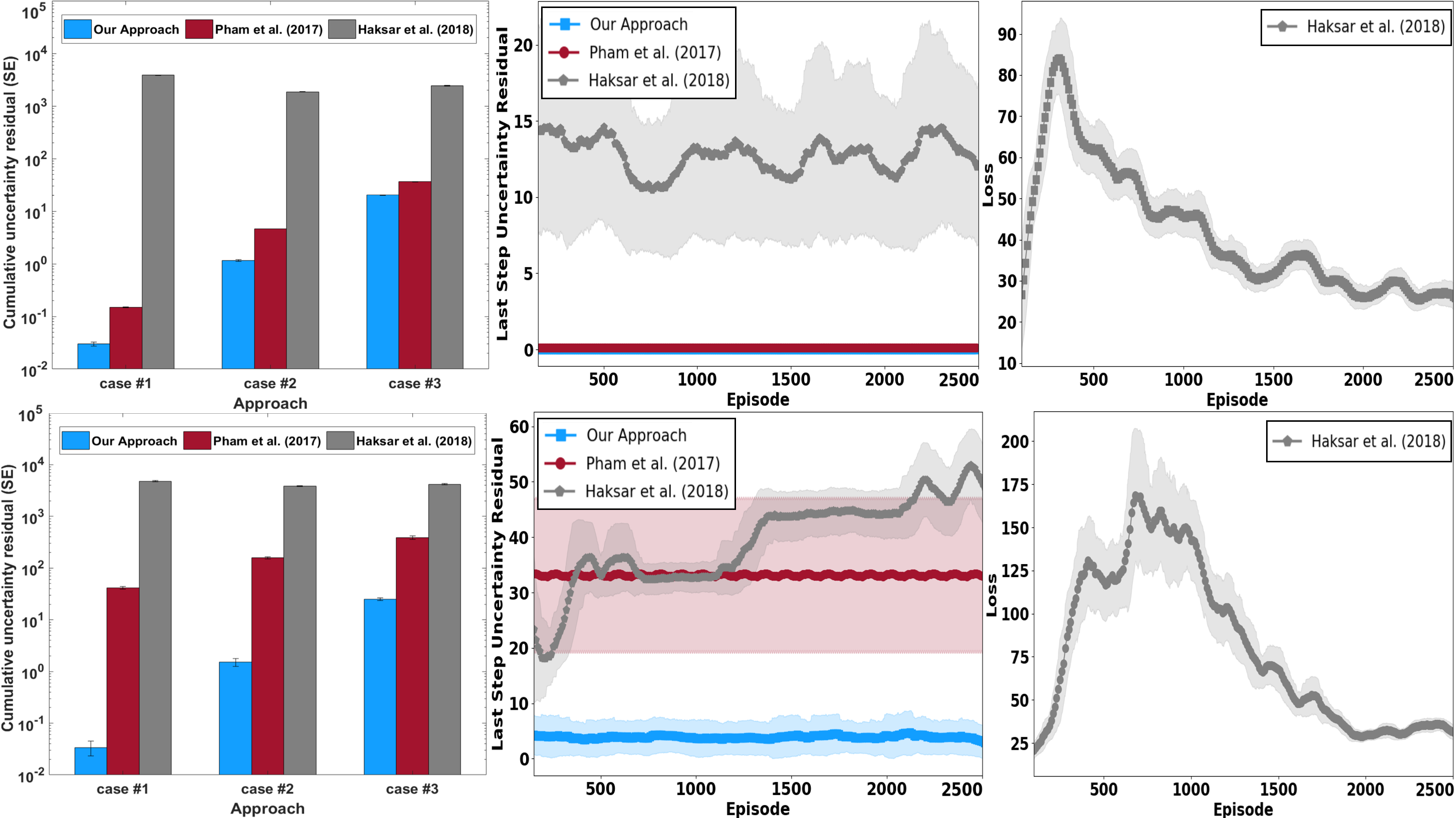}
	\caption{This figure presents a comparison of our predictive coordinated controller to prior works for both our fire environment (bottom row) and that of~\cite{haksar2018distributed} (top row). First column figures show the cumulative uncertainty residual during 100 simulation time steps (10 trials). Middle column figures show the uncertainty residual at the end of last simulation step over the course of 2500 episodes of 100 time-step simulation. The third column figures show the RL agent's loss over the course of training.}
	\label{fig:results_all}
	\vspace*{-0.5cm}
\end{figure*}

\noindent Furthermore, we also evaluated our method by assessing its performance in comparison with two state-of-the-art benchmark studies (\cite{pham2017distributed} and \cite{haksar2018distributed}) for wildfire coverage. In our simulations, we tested all three algorithms in two different fire environments: (1) our fire environment where one area of fire including 40 randomly-placed firespots was initialized and a total of five randomly positioned \textcolor{black}{UAV}s within a distant location were assigned to cover the fire area within a 500-by-500 terrain. (2) the fire environment utilized in~\cite{haksar2018distributed} where a total of 16 firespots all within a 4-by-4 square in the center of a 50-by-50 terrain were initialized and a total of 10 UAVs positioned around this initial square were assigned to cover the fire area. For both environments, \textcolor{black}{UAV}s were required to cover the entire fire map as much as possible, while maintaining an altitude that was both safe and conducive to high-quality imaging. Moreover, the fire model parameters, $R_t$, $U_t$ and $ \theta_t $, were initialized as in~\cite{pham2017distributed} for comparison. The results for these evaluations are presented in Figure~\ref{fig:results_all} in which top and bottom rows correspond to (1) our fire environment and (2) that of~\cite{haksar2018distributed}, respectively.

\subsection{Discussion}
\label{subsec:discussion}
\noindent We investigated the sensitivity of our algorithm to the fire scenario, number of \textcolor{black}{separate areas to be monitored}, and the number of \textcolor{black}{UAV}s available, as shown in Figure~\ref{fig:humanSafety_module_results_all}. The left-hand figure shows that \textcolor{black}{increasing the number of disjoint areas to be monitored resulted in a rise in the number of \textcolor{black}{UAV}s required to guarantee the performance.} This also held as the wildfire propagation scenario changed and a fire propagated more aggressively. As represented in Figure~\ref{fig:humanSafety_module_results_all} (right-side figure), \textcolor{black}{the UAV team was able to easily cover} the fire map in the case of stationary wildfire, while in the case of a fast-growing fire, more \textcolor{black}{UAV}s were required to cover the fire map efficiently \textcolor{black}{(i.e., with low cumulative measurement uncertainty residual)}.

\textcolor{black}{As shown in Figure~\ref{fig:TUB_All}, the computed upper-bound service times, $T_{UB}$ initially start with a high value and gradually decrease until converging to a reasonable value, as the algorithm determines that more UAVs are required for a guaranteed tracking of the firespots. Note that the initial and the converged values of the $T_{UB}$ times have higher values for wildfire Cases 2 and 3 since the moving and moving-spreading firefronts result in expanded search graphs.}

The comparison between our approach and control-theoretic~\cite{pham2017distributed} and reinforcement learning-based~\cite{haksar2018distributed} benchmarks shows that we are able to achieve a $7.5\times$ and $9.0\times$ reduction in error residual for the most challenging cases, as depicted in Figure~\ref{fig:results_all}. For all cases, our approach achieved significantly lower uncertainty residual and cumulative uncertainty in both fire environments. These results demonstrate that our framework not only provides probabilistic bounds on \textcolor{black}{guaranteeing performance}, but also achieves an empirically more-optimal solution for maintaining a tight track on wildfire propagation. Herein, we declare that the main objective for these benchmark comparisons is to evaluate the soundness and feasibility of our coordinated field coverage approach \textcolor{black}{in Section~\ref{subsec:distributedcoverage}} as compared to standard state-of-the-art approaches for this purpose and to evaluate whether our \textit{more-informed} approach can fulfill the expectations by producing comparable and/or better results.

Moreover, we note that the reward-function specification problem and consequently scalability issue in RL-based methods are present, as the RL agent fails to achieve good performance even after convergence (third column from left in Figure~\ref{fig:results_all}). Possible reasons for this include an over- or under-specified reward function from prior work, which emphasized fixed-wing aircraft flight patterns and did not explicitly encourage maximal coverage.

\textcolor{black}{We note that, as discussed in Section~\ref{subsubsec:URR}, an unsatisfied URR bound means that the uncertainty residual \textcolor{black}{may grow with each tour of the UAVs}, which is an indicator that the \textcolor{black}{UAVs may not be} capable of monitoring the targets without losing the track quality. Therefore, a URR greater than one only means that the tracking quality cannot be guaranteed in the respective application. However, the strictness of this bound and whether a UAV can continue the tracking task with small growth in the uncertainty (e.g., 10\% growth) depend on the rate of this growth over time and if such growth is acceptable in the respective application.}

Additionally, the URR upper-bound is designed to recruit UAVs into the monitoring team. However, the URR bound also can be readily used for dismissing UAVs in a simple procedure such as: if URR bound is satisfied with $n$ UAVs then check URR with $n-1$ UAVs (graphs and tours must be recalculated with $n-1$) and set $n = n-1$ if URR is still satisfied. Otherwise, $n$ UAVs will remain. \textcolor{black}{Note that, the URR bound is checked by each UAV and for all the firespots in their assigned region. To select a UAV to be dismissed when the URR bound is satisfied with $n-1$ UAVs, we can choose from regions that are close to each other. This can be achieved, for instance, by calculating the relative distances between the region centroids. When a UAV is dismissed in this way, the reclustering process of the regions is performed through a centralized clustering step such as applying a $k$-means with $n-1$ desired clusters. We then assign UAVs to their new regions.}

\begin{figure}[t!]
	\centering
	\includegraphics[width=\columnwidth]{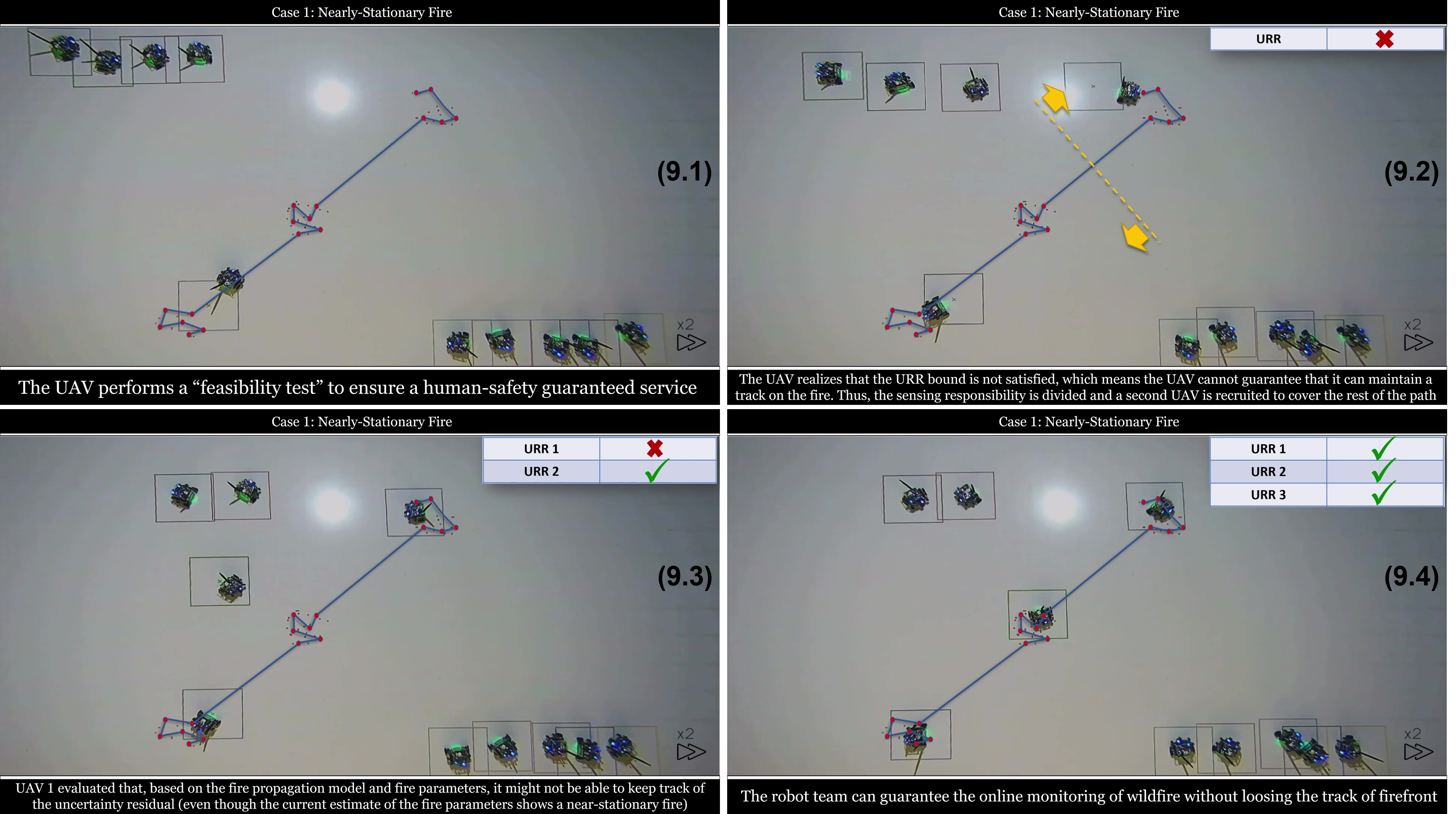}
	\caption{This figure presents example demonstrations of the proposed coordinated planning algorithm for human safety, implemented on physical robots. In Figure (9.4) all URRs are satisfied and the algorithm determines that a team of three robots can collectively provide the estimated information for the three areas without losing the track of any spot. Link to video: \texttt{\url{https://youtu.be/zTR07cKlwRw}}.}
	\label{fig:experiment_results_all1}
	\vspace*{-0.5cm}
\end{figure}

\section{Demonstration: Multi-robot Testbed}
\label{sec:experimentalresults}
\begin{figure}[t!]
	\centering
	\includegraphics[width=\columnwidth]{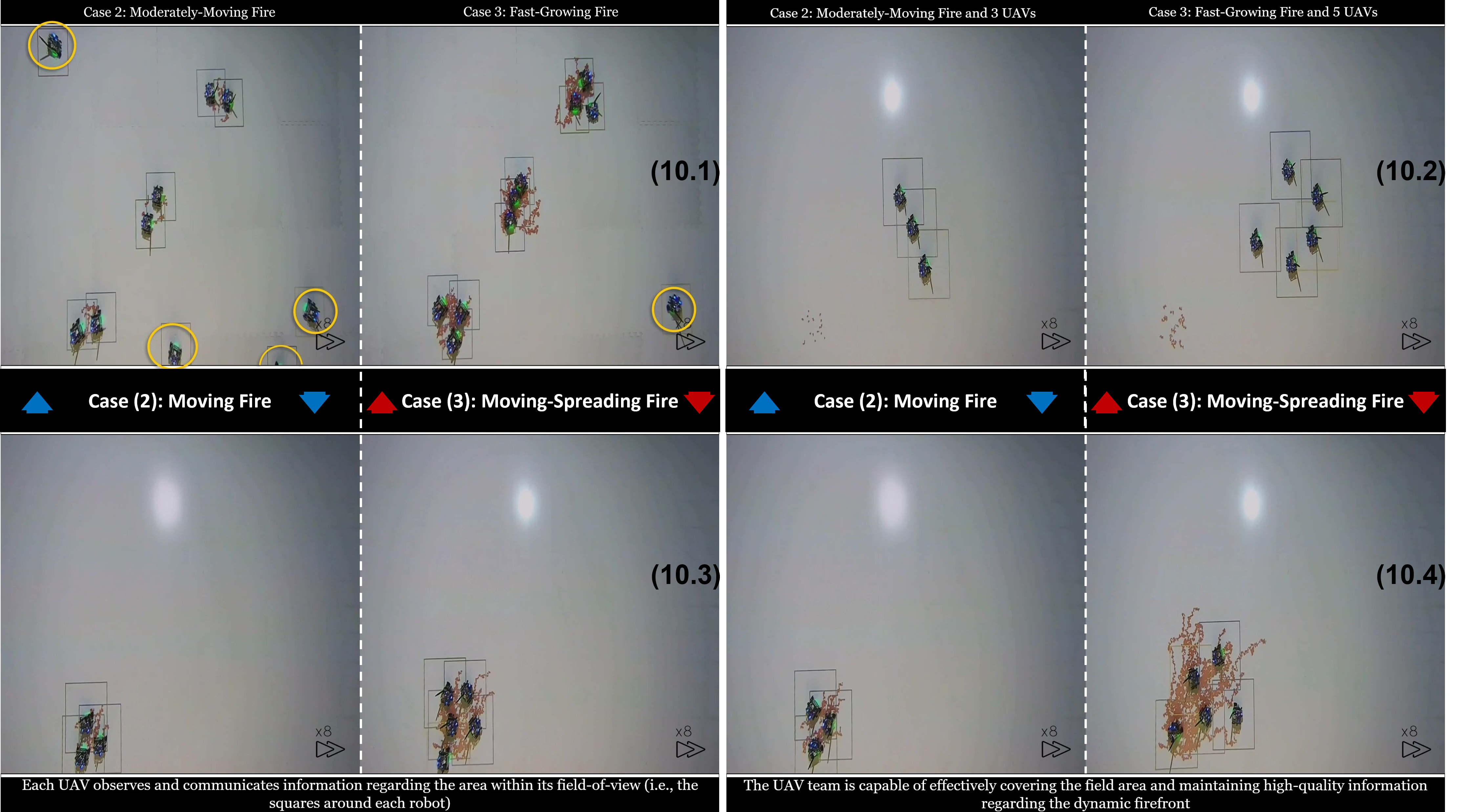}
	\caption{This figure presents four sample demonstrations of our experiments for the proposed coordinated field coverage algorithm, implemented on physical robots. Each sub-figure 10.1 -- 10.4 includes two images: (1) moderate moving fire (left-side) and (2) fast moving-spreading fire (right-side). Figure 10.1 represents the initial step in the algorithm in which \textit{unallocated} robots are selected. Figure 10.4 shows that the robot team can successfully cover both wildfire cases and monitor the dynamic spots,  after ten minutes of running the algorithm. Link to video: \texttt{\url{https://youtu.be/zTR07cKlwRw}}}
	\label{fig:experiment_results_all2}
	\vspace*{-0.5cm}
\end{figure}

In order to account for vehicle dynamics and motion constraints, we implemented and tested \textcolor{black}{our coordinated planning and distributed field coverage} modules in the Robotarium multi-agent robotic platform~\cite{pickem2017robotarium}, at the Georgia Institute of Technology. The simulated growing wildfires using the introduced FARSITE~\cite{finney1998farsite} model are projected on the arena as the regions to be covered by the robot team. The fire simulation parameter setup for our experiments are similar to the empirical evaluation case. In Figures~\ref{fig:experiment_results_all1} and~\ref{fig:experiment_results_all2}, the sub-figures (1)-(4) illustrate initial to final robots standings in our experiments for (1) coordinated planning algorithm \textcolor{black}{for tracking firefronts within specified separate areas} and (2) the coordinated, \textcolor{black}{distributed field coverage} method, respectively. The boxes around each robot shows their respective altitude-dependent FOV. All experiments are repeated for the three wildfire cases and video recordings of these experiments can be found on \texttt{\url{https://youtu.be/zTR07cKlwRw}}.

For the coordinated planning algorithm with URR condition for \textcolor{black}{performance guarantee}, we specified ten robots and tested the feasibility of the algorithm for all three wildfire cases. In the first case, only three robots were needed to \textcolor{black}{guarantee the quality of service} at all times, while this number was six and nine for the second and third scenarios, respectively. Figure~\ref{fig:experiment_results_all1} presents example demonstrations of these experiments. Figure 9.1 shows the beginning of the experiment where a single robot is required to track and monitor three distant areas of fire (\textcolor{black}{i.e.}, simulated firespots projected on the arena). The robot path is also shown. In Figure 9.2 the robot determines it cannot \textcolor{black}{guarantee to provide real-time} information \textcolor{black}{about} all three areas since the URR (shown on top left) is not satisfied and thus, the map is partitioned and another robot is summoned. The same process is repeated in Figures 9.3 -- 9.4 until all URRs are satisfied and the three robots can collectively provide the estimated information for the three areas without losing the track of any spot. Video recordings of these experiments for all three wildfire scenarios can be found on the \textit{first} part of the provided supplementary video.

Next, we tested the coverage performance using five and three robots, again for all three wildfire scenarios. Figure~\ref{fig:experiment_results_all2} presents four sample demonstrations of our experiments for the proposed coordinated field coverage algorithm, implemented on physical robots. Each sub-figure 10.1 -- 10.4 includes two images: (1) moderate moving fire (left-side) and (2) fast moving-spreading fire (right-side). Figure 10.1 represents the initial step in the algorithm in which \textit{unallocated} robots are selected. Figures 10.2 and 10.3 show the team of robots swarming towards the wildfire area and beginning the surveillance. Figure 10.4 shows that the robot team can successfully cover both wildfire cases and monitor the dynamic spots, after ten minutes of running the algorithm. Video recordings of these experiments can be found on the \textit{second} part of the provided supplementary video.

\section{\textcolor{black}{Limitations and Future Work}}
\label{sec:Limitations}
\noindent \textcolor{black}{As stated in Section~\ref{subsec:errorpropagation}, our coordinated planning algorithm is model-based. In our framework, UAVs quantify their estimation uncertainty by explicitly inserting the inferred model parameters from the environment into the model and following a nonlinear uncertainty propagation law. According to our experiments, a mismatch between the actual model used by the algorithm and the ground truth model can lead to significantly increasing the actual measurement uncertainty residuals and consequently an unreliable coverage plan. While this poses a limitation on the framework, in such cases, the best practice is to use adaptive estimation methods Adaptive EKF (AEKF) where the covariances can change adaptively. Nevertheless, the results of such methods can still be unreliable. }

\textcolor{black}{We performed an experiment to evaluate the performance of the AEKF in our framework for cases when there exists a mismatch between the process model and the actual fire propagation model. To impose a model mismatch, we altered the $LB(U_t)$ equation in the actual FARSITE model (introduced in Section~\ref{subsec:simplifiedfarsite}) to be $LB_{new} = \mathcal{X}_1.LB(U_t) + \mathcal{X}_2$, where $\mathcal{X}_1\sim\mathcal{N}(10, 3)$ and $\mathcal{X}_2\sim\mathcal{N}(0, 3)$. This alteration would also alter the values of $GB(U_t)$ and $C(R_t, U_t)$. However, we did not revise the derivatives in the process Jacobian matrix, $F_t$, to account for this alteration. The rest of the environment parameters were \textcolor{black}{exactly the same as in} Section~\ref{subsubsec:EvaluatingtheAlgorithms}. We repeated the simulation for ten separate trials and recorded the cumulative uncertainties over each run. As a result, our experiments demonstrate that the described mismatch between the actual FARSITE model and the process model in the adopted AEKF, on average, caused $1.48\pm0.4$ times higher cumulative uncertainties across wildfire scenarios, as compared to the case without a model mismatch. This increase in uncertainty residual demonstrates that even an adaptive estimation method such as AEKF cannot completely compensate for a model mismatch.}

We discuss that in our approach both UAVs or a ground \textit{control room} can be in charge of running the algorithm. By assuming local communication between neighboring UAVs, without losing generality, we can presume either case to be practical, assuming required hardware and computational resources are available. By assuming an on-board computer, each UAV can perform the URR bound check in Equation~\ref{eq:URR1} and communicate its local belief of its assigned path to a ground control room, which then recruits additional UAVs to the task. We note that distributed communications could result in bottlenecks in our coordinated field coverage method; however, we rely on the rich literature on algorithms available to manage wireless communication grids to address this problem (see~\cite{danoy2015connectivity,li2008robot}).

\textcolor{black}{In Section~\ref{subsec:T_UB}, we made a worst-case scenario assumption regarding the independence of firespots and uniformity of fastest and universal velocity for all firespots. While these assumptions are made to generalize the applicability of the approach, in future work, the independence assumption can be relaxed for the wildfire monitoring application, given that an accurate model of the correlations between firespots are in hand. Accordingly, the calculated upper-bound service time, $T_{UB}$, will become smaller and closer to the actual service time, $T^*$, making the generated monitoring plans more optimal.}

\section{Conclusion}
\label{sec:discussion}
\noindent We have introduced a novel analytical measurement-residual bound on fire propagation uncertainty, allowing high-quality planning for real-time wildfire monitoring and tracking, while also providing a probabilistic guarantee on \textcolor{black}{the quality of service}. Our approach outperformed prior work for distributed control of UAVs for wildfire tracking, as well as a reinforcement learning baseline. Our quantitative evaluations validate the performance of our method accumulating $7.5\times$ and $9.0\times$ less error residual than the model based benchmark~\cite{pham2017distributed} and the learning-based benchmark~\cite{haksar2018distributed} respectively when covering a large aggressive wildfire, over a course of 2500 episodes of 100-steps long simulations. \textcolor{black}{Our method also outperformed the best-performing baseline~\cite{pham2017distributed} in a challenging evolving fire scenario by accumulating $5.8\times$ less uncertainty residual than the baseline over the simulation time and on average per time-step method.} See Section~\ref{sec:results} and~\ref{subsec:discussion} for the evaluations and a discussion on the presented results. Physical implementation of our framework on real robots in a multi-robot testbed demonstrate and validate the feasibility of our approaches.

We note that in this work we generalize by considering large-scale human-defined \textcolor{black}{areas of priority for monitoring and tracking moving targets (such as firespots) and investigating three different scenarios of stationary targets (e.g., the traveling salesman problem and search-and-rescue), moving targets (e.g., border patrol and the tracking of wildlife poachers) and moving-spreading targets (e.g., wildfire monitoring and oil spill surveillance).} We also note that due to the modular design of our framework here, any distributed control algorithm for dynamic field coverage, such as our introduced previous work~\cite{seraj2020coordinated}, can replace the proposed coordinated coverage module in section~\ref{subsec:distributedcoverage} \textcolor{black}{to enable \textit{unallocated} UAVs to monitor the unspecified areas of wildfire}.


\section*{\textcolor{black}{Appendix}}
\subsection*{\textcolor{black}{A.1 Time Independency of the EKF's Measurement Residual}}
\label{subsec:ProofOfTheorem2}
\noindent \textcolor{black}{Our analytical URR bound in Equation~\ref{eq:URR1} depends on the state-estimation measurement residual computed at different time-steps. \textcolor{black}{To maintain control over the measurement uncertainty, we posit that the UAV observers would want the measurement uncertainty residual with respect to a target on the ground not to increase from $t=t_0$ to $t=t_0+kT_{UB}$ for any positive integer constant $k$ if the UAV observes the target from the same relative position. Therefore,} we examine the time-dependency of the propagated error through our EKF formulation. To this end, we follow the mathematical proof and discussions provided in~\cite{seraj2021hierarchical} and \cite{seraj2020coordinated, seraj2019safe}. \textcolor{black}{We state that the measurement uncertainty about the states of a dynamic point $ q_t $, observed by a flying UAV is independent of time and is only a function of distance between the observer and the point. In the following, we mathematically proof this point.}}

\textcolor{black}{\textcolor{black}{First, we present how the uncertainty residual is quantified by an EKF. The total uncertainty residual propagated by EKF is composed of a model and an observation measurement uncertainties, both of which} follow the general nonlinear uncertainty propagation law, shown in Equations~\ref{eq:ekfModelUncertainty11}-\ref{eq:ekfObservationUncertainty11}, where $ \Sigma_{t|t-1} $ is the predicted covariance estimate, $ \Lambda_{t|t} $ is the innovation (or residual) covariance, $ F_t $ and $ H_t $ are the process and observation Jacobian matrices, and $ Q_t $ and $ \Gamma_t $ are the process and observation noise covariances, respectively.
\begin{align}
           \Sigma_{t|t-1} &= F_t\Sigma_{t-1|t-1}F_t^T + Q_t \label{eq:ekfModelUncertainty11} \\
           \Lambda_{t|t} &= H_t\Sigma_{t|t-1}H_t^T + \Gamma_t \label{eq:ekfObservationUncertainty11}
\end{align}Considering Equations~\ref{eq:ekfModelUncertainty11}-\ref{eq:ekfObservationUncertainty11}, the gradients in the \textcolor{black}{process, $ F_t $, and observation, $ H_t $, Jacobian matrices are responsible for alterations in the uncertainty values}. \textcolor{black}{To compute these gradients, we calculate the derivatives of fire's propagation model, $ \mathcal{M}_t $, and UAV's observation model, $ \mathcal{O}_t $, with respect to the state variables. As discussed in Section~\ref{subsec:errorpropagation} and considering the introduced state vectors, we first derive the process and observation Jacobian matrices ($ F_t $ and $ H_t $) as follows in Equations~\ref{eq:statetransitionJacob}-\ref{eq:observationjacob}, respectively. In Equations~\ref{eq:statetransitionJacob}-\ref{eq:observationjacob},} $ t^\prime = t-1 $.}
\textcolor{black}{\begin{align}
\label{eq:statetransitionJacob}
\left. \frac{\partial \mathcal{M}_t}{\partial \mathcal{S}_i}\right|_{\hat{\Theta}_{t|t'}} = \begin{blockarray}{ccccccccc}
& q_{t^\prime}^x & q_{t^\prime}^y & p_{t^\prime}^x & p_{t^\prime}^y & p_{t^\prime}^z & R_{t^\prime} & U_{t^\prime} & \theta_{t^\prime} \\
\begin{block}{c(cccccccc)}
q_t^x & 1 & 0 & 0 & 0 & 0 & \frac{\partial q_{t}^x}{\partial R_{t^\prime}}  & \frac{\partial q_{t}^x}{\partial U_{t^\prime}} & \frac{\partial q_{t}^x}{\partial \theta_{t^\prime}} \\
q_t^y & 0 & 1 & 0 & 0 & 0 & \frac{\partial q_{t}^y}{\partial R_{t^\prime}}  & \frac{\partial q_{t}^y}{\partial U_{t^\prime}} & \frac{\partial q_{t}^y}{\partial \theta_{t^\prime}} \\
p_{t}^x & 0 & 0 & 0 & 0 & 0 & 0 & 0 & 0  \\
p_{t}^y & 0 & 0 & 0 & 0 & 0 & 0 & 0 & 0  \\
p_{t}^z & 0 & 0 & 0 & 0 & 0 & 0 & 0 & 0  \\
R_t & 0 & 0 & 0 & 0 & 0 & 1 & 0 & 0  \\
U_t & 0 & 0 & 0 & 0 & 0 & 0 & 1 & 0  \\
\theta_t & 0 & 0 & 0 & 0 & 0 & 0 & 0 & 1  \\
\end{block}
\end{blockarray}
\end{align}}
\textcolor{black}{\begin{align}
\label{eq:observationjacob}
\left. \frac{\partial \mathcal{O}_t}{\partial \Phi_i}\right|_{\hat{\Phi}_{t}} = \begin{blockarray}{ccccccccc}
& q_{t}^x & q_{t}^y & p_{t}^x & p_{t}^y & p_{t}^z & R_{t} & U_{t} & \theta_{t} \\
\begin{block}{c(cccccccc)}
\varphi_t^x & \frac{\partial \varphi_{t}^x}{\partial q_t^x} & \frac{\partial \varphi_{t}^x}{\partial q_t^y} & \frac{\partial \varphi_{t}^x}{\partial p_t^x} & \frac{\partial \varphi_{t}^x}{\partial p_t^y}  & \frac{\partial \varphi_{t}^x}{\partial p_t^z} & 0 & 0 & 0 \\     
\varphi_t^y & \frac{\partial \varphi_{t}^y}{\partial q_t^x} & \frac{\partial \varphi_{t}^y}{\partial q_t^y} & \frac{\partial \varphi_{t}^y}{\partial p_t^x} & \frac{\partial \varphi_{t}^y}{\partial p_t^y} 
& \frac{\partial \varphi_{t}^y}{\partial p_t^y} & 0 & 0 & 0 \\
\hat{R}_t & 0 & 0 & 0 & 0 & 0 & 1 & 0 & 0 \\
\hat{U}_t & 0 & 0 & 0& 0 & 0 & 0 & 1 & 0 \\
\hat{\theta}_t & 0 & 0 & 0 & 0 & 0 & 0 & 0 & 1 \\
\end{block}
\end{blockarray}
\end{align}}\textcolor{black}{In Eq.~\ref{eq:statetransitionJacob}-\ref{eq:observationjacob}, we define the process state vector as $ \Vec{\Theta}_t = \left[ q_t^x, q_t^y, p_t^x, p_t^y, p_t^z, R_t, U_t, \theta_t \right]^T $ and $ \Vec{\Phi}_t = \left[\varphi_t^x, \varphi_t^y, \hat{R}_t, \hat{U}_t, \hat{\theta}_t\right]^T $ as the mapping vector. As such, we calculate the partial derivatives in Eq.~\ref{eq:statetransitionJacob} by using Eq.~\ref{eq:firemotionmodel}-\ref{eq:qdot1} and applying the chain-rule to compute the derivatives of $ q_t^x $ and $ q_t^y $ with respect to parameters $ R_{t-1} $, $ U_{t-1} $, and $ \theta_{t-1} $. The partial derivatives are then derived as in Eq.~\ref{eq:qJacobs1}-\ref{eq:qJacobs2},} \textcolor{black}{where $ \mathcal{D}(\theta) $ is $ \sin\theta $ and $ \cos\theta $ for X and Y axis, respectively.}
\textcolor{black}{\begin{align}
   \frac{\partial q_{t}}{\partial \theta_{t-1}} =& ~C(R_t, U_t)\frac{\partial\mathcal{D}(\theta)}{\partial\theta}\delta t \label{eq:qJacobs1}\\
   \frac{\partial q_{t}}{\partial R_{t-1}} =& \left(1-\frac{LB(U_t)}{LB(U_t) + \sqrt{GB(U_t)}}\right)\mathcal{D}(\theta)\delta t \\
   \frac{\partial q_{t}}{\partial U_{t-1}} =& \frac{R_{t'}\bigg(LB(U_{t'})\frac{\partial GB(U_{t'})}{\partial U_{t'}} - GB(U_{t'})\frac{\partial LB(U_{t'})}{\partial U_{t'}}\bigg)}{\left(LB(U_{t'})+\sqrt{GB(U_{t'})}\right)^2}\mathcal{D}(\theta)\delta t \label{eq:qJacobs2}
\end{align}}\textcolor{black}{\textcolor{black}{To compute the partial derivatives in the observation Jacobian matrix in Equation~\ref{eq:observationjacob}, we first need to derive the relation between the angle parameters, $ \varphi_t^x $ and $ \varphi_t^y $, and the UAV pose.} The angle parameters contain information regarding both firefront location $ [q_t^x, q_t^y] $ and UAV coordinates $ [p_t^x, p_t^y, p_t^z] $. According to Fig.~\ref{fig:ObsMdl}, by projecting the looking vector of UAV to planar coordinates, the angle parameters are \textcolor{black}{calculated as shown in Equations\ref{eq:angleParamXX}-\ref{eq:angleParamYY}} for X and Y axes respectively, where $ q_t = [q_t^x, q_t^y] $ and $ p_t = [p_t^x, p_t^y] $.}
\textcolor{black}{\begin{align}
    \label{eq:angleParamXX}
    \varphi_t^x &= \tan^{-1}\left(\frac{p_t^z}{\|q_t-p_t\|}\right) \\
    \varphi_t^y &= \tan^{-1}\left(\frac{\|q_t-p_t\|}{p_t^z}\right) 
\label{eq:angleParamYY}
\end{align}}\textcolor{black}{The partial derivatives in the observation Jacobian matrix $ H_t $ for \textit{X}-axis, presented in Eq.~\ref{eq:observationjacob}, are derived as in Eq.~\ref{eq:Hjacobsxaa}-\ref{eq:Hjacobsxx} and for \textit{Y}-axis derivatives, we can derive as in Eq.~\ref{eq:Hjacobsyaa}-\ref{eq:Hjacobsyy}.}
\textcolor{black}{\begin{align}
\label{eq:Hjacobsxaa}
\mathlarger{\mathlarger{\nabla}}_{q_t}\varphi^x_t = & \frac{1}{1+\left(\frac{p_t^z}{\|q_t-p_t\|}\right)^2}\left(\frac{-p_t^z\left(q_t-p_t\right)}{\|q_t-p_t\|^3}\right)=\left[\frac{\partial \varphi_{t}^x}{\partial q_t^x}, \frac{\partial \varphi_{t}^x}{\partial q_t^y}\right]\\
\mathlarger{\mathlarger{\nabla}}_{p_t}\varphi^x_t = & \frac{1}{1+\left(\frac{p_t^z}{\|q_t-p_t\|}\right)^2}\left(\frac{p_t^z\left(q_t-p_t\right)}{\|q_t-p_t\|^3}\right)=\left[\frac{\partial \varphi_{t}^x}{\partial p_t^x}, \frac{\partial \varphi_{t}^x}{\partial p_t^y}\right]\\
\frac{\partial \varphi_{t}^x}{\partial p_t^z} = & \frac{1}{1+\left(\frac{p_t^z}{\|q_t-p_t\|}\right)^2}\left(\frac{1}{\|q_t-p_t\|}\right) \label{eq:Hjacobsxx}
\end{align}}
\textcolor{black}{\begin{align}
\label{eq:Hjacobsyaa}
\mathlarger{\mathlarger{\nabla}}_{q_t}\varphi^y_t = & \frac{1}{1+\left(\frac{\|q_t-p_t\|}{p_t^z}\right)^2}\left(\frac{\left(q_t-p_t\right)}{p_t^z\|q_t-p_t\|}\right)=\left[\frac{\partial \varphi_{t}^y}{\partial q_t^x}, \frac{\partial \varphi_{t}^y}{\partial q_t^y}\right]\\
\mathlarger{\mathlarger{\nabla}}_{p_t}\varphi^y_t = & \frac{1}{1+\left(\frac{\|q_t-p_t\|}{p_t^z}\right)^2}\left(\frac{-\left(q_t-p_t\right)}{p_t^z\|q_t-p_t\|}\right)=\left[\frac{\partial \varphi_{t}^y}{\partial p_t^x}, \frac{\partial \varphi_{t}^y}{\partial p_t^y}\right]\\
\frac{\partial \varphi_{t}^y}{\partial p_t^z} = & \frac{1}{1+\left(\frac{\|q_t-p_t\|}{p_t^z}\right)^2}\left(\frac{-\|q_t-p_t\|}{(p_t^z)^2}\right) \label{eq:Hjacobsyy}
\end{align}}\textcolor{black}{\textcolor{black}{Now, considering EKF's covariance propagation Equations} in Equations~\ref{eq:ekfModelUncertainty11}-\ref{eq:ekfObservationUncertainty11} as well as the gradients in process Jacobian matrix $ F_t $ as calculated \textcolor{black}{in Eq.~\ref{eq:qJacobs1}-\ref{eq:qJacobs2}, we can see that }the gradients in process Jacobian matrix are only functions of fire propagation model parameters (\textcolor{black}{e.g., the FARSITE model in this case}) such as fuel coefficient, $ R_t $ \textcolor{black}{and wind velocity and direction, $ U_t $ and $ \theta_t $.} Consequently, while these parameters do not vary significantly with time, the uncertainty drop due to process model is time-invariant. We note that FARSITE~\cite{finney1998farsite} assumes locality in time (i.e., within seconds or few minutes), making the assumption of time-invariant fire parameters fairly acceptable~\cite{delamatar2013downloading}. \textcolor{black}{Moreover}, the gradients in the observation Jacobian \textcolor{black}{matrix, Eq.~\ref{eq:Hjacobsxaa}-\ref{eq:Hjacobsyy}, are only} functions of the Euclidean distance between the \textcolor{black}{UAV pose and firespot coordinates}. \textcolor{black}{We also know that, since} at the time of visiting a firespot the planar displacement between UAV and fire locations are \textcolor{black}{approximately} zero and the only distance between the two \textcolor{black}{equals to} the UAV altitude. Accordingly, both $ F_t $ and $ H_t $ are \textcolor{black}{locally} time-invariant and the total \textcolor{black}{measurement} uncertainty residual variations between two different time-steps (\textcolor{black}{e.g., $t=t_0$ and $t=t_0+kT_{UB}$}) is not a function of time \textcolor{black}{and is only a function of the UAV observer's altitude}.}




\bibliographystyle{IEEEtran}
\bibliography{IEEEabrv,ICRA20}


\end{document}